\documentclass[10pt,conference]{IEEEtran}
\IEEEoverridecommandlockouts

\usepackage[T1]{fontenc}
\usepackage{newtxtext,newtxmath}
\usepackage{amsmath,booktabs,tabularx,array,graphicx,microtype,xcolor,enumitem,xspace}
\usepackage[colorlinks=true,linkcolor=blue!50!black,citecolor=blue!55!black,urlcolor=blue!55!black]{hyperref}
\usepackage[backend=biber,style=authoryear,sorting=nyt,labelnumber=true,maxcitenames=1,uniquelist=false,date=year,maxbibnames=999,giveninits=false,uniquename=false,backref=true,doi=true,url=true,isbn=false]{biblatex}
\DeclareNameAlias{sortname}{given-family}
\DeclareNameAlias{author}{given-family}

\hypersetup{
  unicode=true,
  pdftitle={RoboNoid: Scaling Humanoid Intelligence through Agentic Composition and Adaptation},
  pdfauthor={Yu Bai, Ziyi Bai, Ming-Ming Yu, Chaojie Li, Chaoyi Xu, Jintao Li, Siqi Li, Boyu Li, Xinrun Xu, Shuyi Lin, Heng Wang, Zhiwei Yu, B\"orje F. Karlsson},
  pdfsubject={Humanoid intelligence and autonomous embodied agents},
  pdfkeywords={Humanoid robotics, embodied AI, autonomous agents, RGB-centric learning, agentic composition, AI agents, agentic systems, adaptation, situated experience, scalability},
  pdflang={en-US}
}

\renewbibmacro*{in:}{%
  \ifboolexpr{test {\ifentrytype{article}} or test {\ifentrytype{unpublished}}}
    {}{\printtext{\bibstring{in}\intitlepunct}}}
    
\DeclareFieldFormat{labelnumberwidth}{\mkbibbrackets{#1}}
\defbibenvironment{bibliography}
  {\list
     {\printtext[labelnumberwidth]{\printfield{labelprefix}\printfield{labelnumber}}}
     {\setlength{\labelwidth}{\labelnumberwidth}%
      \setlength{\leftmargin}{\labelwidth}%
      \setlength{\labelsep}{\biblabelsep}%
      \addtolength{\leftmargin}{\labelsep}%
      \setlength{\itemsep}{\bibitemsep}%
      \setlength{\parsep}{\bibparsep}}%
   }
  {\endlist}
  {\item}
  
\newbibmacro*{robo:backrefs}{%
  \iflistundef{pageref}{}{%
    \setunit{\addperiod\par\nobreak}%
    \printtext{{\footnotesize\color{blue!55!black}%
      \ensuremath{\hookleftarrow}\,\printlist[pageref][-\value{listtotal}]{pageref}}}}}
\renewbibmacro*{pageref}{\usebibmacro{robo:backrefs}}
\renewbibmacro*{setpageref}{\usebibmacro{robo:backrefs}}
\newcommand{\robonoid}{RoboNoid\xspace}

\newcommand{\AuthorMarks}[1]{\if\relax\detokenize{#1}\relax\else\textsuperscript{\normalfont #1}\fi}
\newcommand{\RoboAuthors}{Yu Bai\AuthorMarks{1}, Ziyi Bai\AuthorMarks{1}, Ming-Ming Yu\AuthorMarks{2*}, Chaojie Li\AuthorMarks{3*}, Chaoyi Xu\AuthorMarks{1}, Jintao Li\AuthorMarks{4*}, Siqi Li\AuthorMarks{4*},\\ Boyu Li\AuthorMarks{1*}, Xinrun Xu\AuthorMarks{4*}, Shuyi Lin\AuthorMarks{3*}, Heng Wang\AuthorMarks{1}, Zhiwei Yu\AuthorMarks{1}, B\"orje F. Karlsson\AuthorMarks{1\textdagger}}

\newcommand{\RoboAffiliations}{Beijing Academy of Artificial Intelligence (BAAI)}
\newcommand{\MarkerA}{Beijing Academy of Artificial Intelligence (BAAI)}
\newcommand{\MarkerB}{Beihang University}
\newcommand{\MarkerC}{Chinese University of Hong Kong, Shenzhen}
\newcommand{\MarkerD}{University of the Chinese Academy of Sciences}
\newcommand{\MarkerE}{Worker performed during author's internship at BAAI}
\newcommand{\MarkerH}{Corresponding author: (borje.karlsson@tellarin.com)}
\newcommand{\AuthorFootnotes}{\begingroup\renewcommand{\thefootnote}{}\footnotetext{\textsuperscript{1}\MarkerA; \textsuperscript{2}\MarkerB; \textsuperscript{3}\MarkerC; \textsuperscript{4}\MarkerD; \textsuperscript{*}\MarkerE;  \textsuperscript{\textdagger}\MarkerH.}\endgroup}

\AddToHook{shipout/foreground}{%
  \ifnum\value{page}>1\relax
    \begingroup
    \setlength{\unitlength}{1bp}%
    \put(56.35,-29){\makebox[496.74bp][l]{\normalfont\fontsize{11}{13}\selectfont\itshape RoboNoid Technical Report}}%
    \put(56.35,-36){\rule{496.74bp}{0.39bp}}%
    \endgroup
  \fi
}

\begin{document}
\pagestyle{plain}

\twocolumn[{\begin{center}

\includegraphics[width=0.98\textwidth]{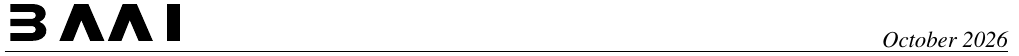}\\[0.55em]
{\fontsize{23}{27}\selectfont RoboNoid: Scaling Humanoid Intelligence\\ through Agentic Composition and Adaptation}\\[0.65em]
{\fontsize{11}{14}\selectfont\RoboAuthors}\\[0.3em]
{\fontsize{10}{12}\selectfont\RoboAffiliations}\\[2.65em]
\refstepcounter{figure}\label{fig:capabilities}
\includegraphics[width=0.83\textwidth]{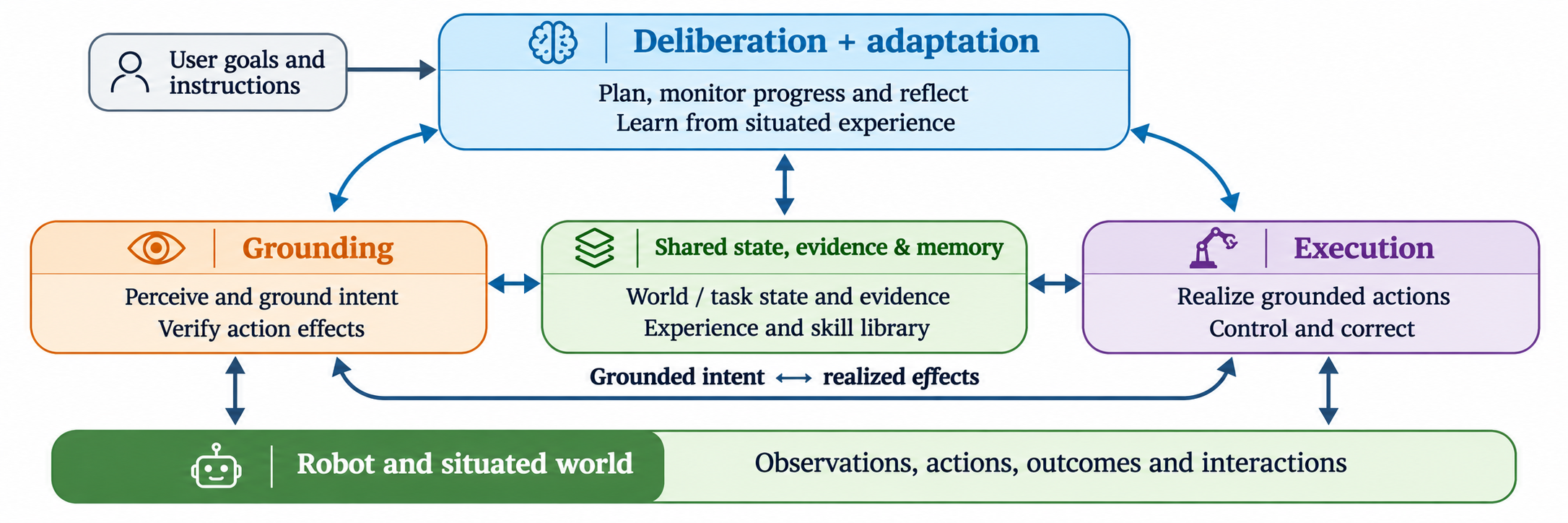}\\[0.4em]
\parbox{0.98\textwidth}{\small \textbf{Fig. \thefigure: \robonoid overview.} 
Interacting agentic loops coordinate deliberation and adaptation (System 2), spatial grounding and verification (System 1), and physical realization and control (System 0). Shared state, evidence, and memory support coordination and improvement through situated experience across different abstractions and timescales. The capability boundaries are not fixed model boundaries.}
\end{center}\vspace{1.45em}}]


\thispagestyle{plain}

\AuthorFootnotes

\begin{abstract}
Despite considerable advances in robotics and foundation models, there is still limited understanding of what data, representations, and architectures are necessary for general-purpose humanoid intelligence. Addressing this challenge requires studying humanoids as autonomous systems operating in realistic settings, rather than merely as collections of individually learned tasks. RoboNoid investigates scalable humanoid intelligence that meets the demands of dynamic real-world environments. Our approach is grounded in two complementary principles: reducing the constraints on data acquisition and reducing the reliance on precision and predictability during execution. The former motivates an RGB-centric approach to learning, making use of visual data that can be acquired at scale across diverse sources and embodiments. The latter calls for agents capable of adapting not only to unfamiliar environments and tasks, but also to noisy observations, imperfect execution, and their own mistakes. Situated experience is essential both to improving individual capabilities and to understanding how they should work together. To this end, we propose a composable agent architecture that allows capabilities to be independently developed, jointly evaluated, and progressively co-evolved. Realistic end-to-end scenarios bringing together locomotion, manipulation, physical reasoning, and human interaction expose challenges that are difficult to anticipate through isolated, task-specific experiments. This report presents the resulting architecture, empirical findings from complementary investigations, and open challenges that inform its evolution.
\end{abstract}


\section{Introduction}
A humanoid can have a reliable walking policy, a capable grasping policy, and a foundation model that understands user instructions, yet probably still fail to bring a requested object to the requesting person. The obstacle may be an unfavourable stance left by navigation, a changed object location, an occluded view, or a mistaken belief that a previous action succeeded. Such failures are not necessarily solved by improving any one component on its own. They emerge from the interaction of capabilities and the environment, and cross task boundaries. More importantly, they show us what information and capabilities a system actually lacks.

The data problem is equally unsettled. Human environments contain a great deal of visual evidence about objects, action possibilities, conventions, and consequences. Robot demonstrations are valuable because they reveal the feasibility and dynamics of actions for a particular embodiment, but they need not be the sole source of supervision for everything a humanoid learns. An approach that requires synchronized, accurately calibrated, embodiment-specific trajectories for all knowledge imposes a substantial constraint on scale. Moreover, depending on extra cameras placed on the environment or in different positions in a robot, or additional sensors, reduce the possibility of reuse of existing data and make models more complex. Likewise, an approach that depends on perfect localization, a precise 3D world model, or high accuracy execution at every step is unlikely to retain that performance outside controlled conditions.

Our position is that such requirements should be reduced where possible, and compensate for the resulting uncertainty through situated adaptation. This is not an argument for ignoring physical constraints or for replacing control with a language model. It is an argument for making the system responsible for observing and correcting its own progress. We use RGB video as the main scalable observation and learning modality, retain proprioceptive and other signals only when they are necessary for feasible and safe execution, and study the composition of capabilities in realistic tasks. The cost of acquiring the next useful capability and the ability to reuse it under changed conditions are as important as the accuracy of a particular model. Additional sensors and data modalities can always be considered to try and further improve system performance, if available.

Through RoboNoid, we investigate how humanoid intelligence can be made scalable while meeting the adaptive requirements of dynamic real-world environments. We propose an agentic architecture that addresses several of the challenges involved in realizing general-purpose humanoid intelligence, while providing a concrete basis for end-to-end experimentation and further capability development. Complementary research explores challenges that remain insufficiently addressed, alternative realizations of capabilities, and mechanisms through which the system can improve from situated experience. Their findings inform the progressive evolution of the architecture. Our contribution is thus threefold: a research approach to scalable and adaptive humanoid intelligence, empirical findings across complementary capabilities and integrated scenarios, and an evolving architecture that supports their composition, experimentation, and co-evolution. 

\robonoid builds on our work in long-horizon agents \parencite{cradle}, humanoid egocentric grounding \parencite{being0,egoactor}, layout understanding \parencite{ranger}, robot-state-aware planning and cross-embodiment action generation \parencite{proprio, xdiffvla}, and experience-driven evolution \parencite{mart, practice}. These studies motivate an agent architecture that permits specialization and composition without committing to permanently distinct models. Figure~\ref{fig:capabilities} summarizes its architecture design.

The three principles are interdependent. Less restrictive data acquisition increases the diversity of available supervision, but also leaves uncertainty about embodiment and action consequences. Adaptation lets the agent resolve some of that uncertainty during execution. Composition lets us investigate which capability or interface must improve, rather than presuppose the data and model architecture required to learn the whole system. Situated experience connects these choices: it exposes the assumptions that fail and provides evidence for their revision.


\setcounter{dbltopnumber}{3}
\renewcommand{\dbltopfraction}{0.90}
\renewcommand{\textfraction}{0.08}

\section{Humanoid intelligence as a systems problem}
\label{sec:running}

This report focuses on the research approach, architecture, and representative examples of embodied behaviour. Detailed experimental protocols and quantitative comparisons are presented in the cited component studies; the examples discussed here illustrate capabilities and system-level requirements.

Consider a request to prepare a coffee and bring it to a person in another room. This requires more than a collection of successful actions: a cup must be reachable from the stance chosen by navigation (which itself requires localization, understanding environment layout, and negotiating obstacles), the appliance must be put in the required mode, the drink must remain secure during transport, and the recipient must be identified. A potentially useful decision at one stage can make the next stage infeasible. The desired effects and the evidence for them, rather than the sequence of skill calls, define task completion. Feedback-based robotic planning, as in Inner Monologue~\parencite{inner}, similarly highlights the dependence of subsequent decisions on observed outcomes.

Humanoid robots make the coordination problem especially visible. Walking, reaching, carrying and interacting with an object all affect the configuration and balance of the robot. A wheeled platform or a simple gripper can be useful for particular tasks, but may hide dependencies that matter when an agent must coordinate the whole body. We therefore use realistic humanoid scenarios as a testbed for the design of the agent, for the collection of situated data, and for the study of generalization. The objective is not to maximize the complexity of a demonstration, but to understand which capabilities and interactions are necessary to accomplish it reliably.

\subsection{From general computer control to embodied agents}

Agentic systems have been shown to handle digital games and computer software. For example, our previous work in Cradle~\parencite{cradle}, which investigates general computer control through visual observations and general keyboard and mouse actions. Its agent framework combines information gathering, self-reflection, task inference, skill curation and action planning, supported by episodic and procedural memory. The approach separates deciding what to accomplish from realizing the corresponding actions in a particular environment. It also motivates the longer-term vision of agents that could operate across digital and physical environments, where manipulating a tangible interface and using a digital application may be parts of the same task. The interleaving of reasoning and environmental actions in ReAct~\parencite{react} and the composition of executable robot programs in Code as Policies~\parencite{cap} provide related foundations.

\begin{figure*}[!t]
    \centering
    \includegraphics[width=.96\textwidth]{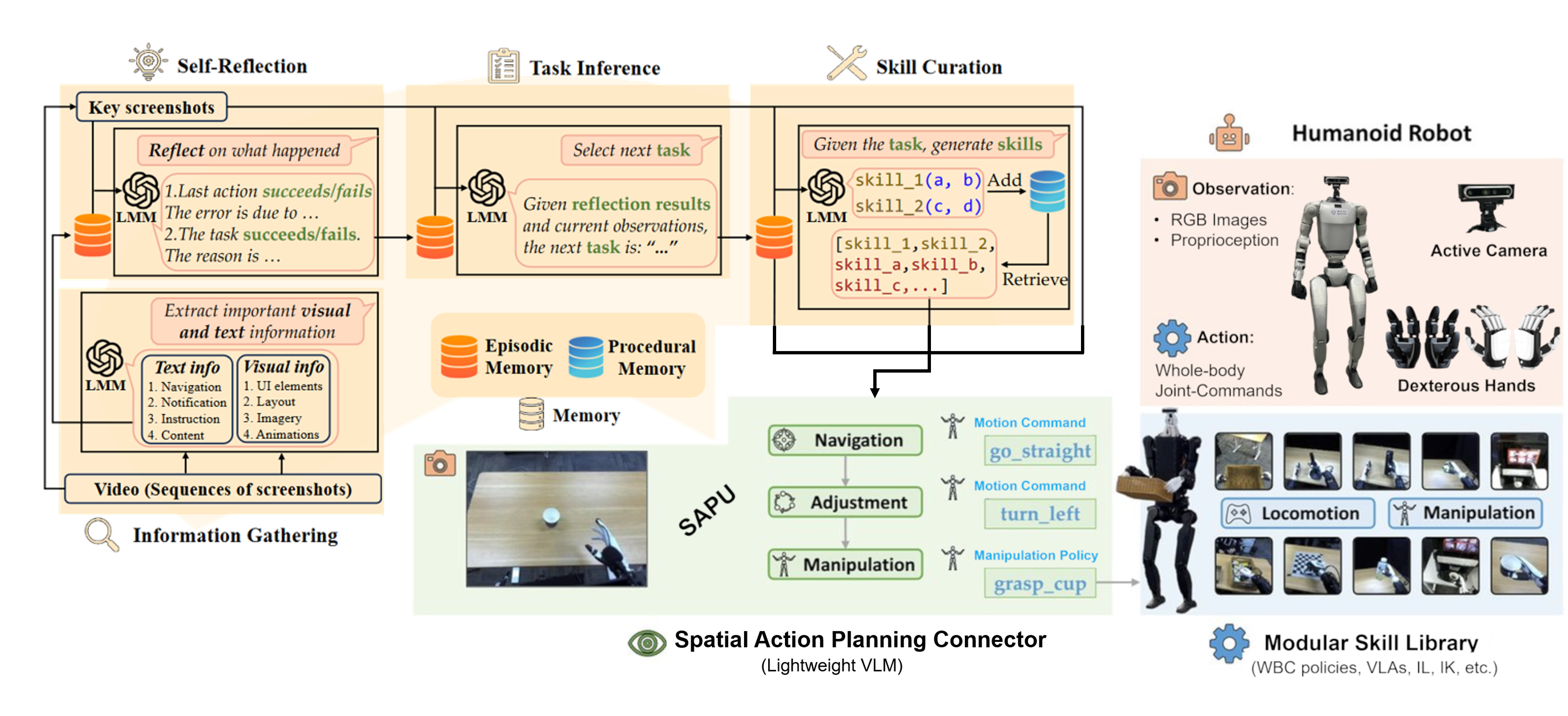}
    \caption{Extending the Cradle agent architecture into humanoid robot control. Information gathering, self-reflection, task inference and skill curation motivate the task-level agent; spatial action planning and the modular skill library connect its decisions to physical execution, highlighting the criticality of real-world grounding.}
    \label{fig:cradle_embodied}
\end{figure*}

A physical agent introduces additional requirements: action feasibility depends on embodiment, balance and contact; sensing and execution change each other's context; and effects may be delayed, uncertain, or irreversible. Figure~\ref{fig:cradle_embodied} illustrates how the agentic ideas of Cradle can be combined with spatial action grounding and a modular robot skill library. It is a conceptual connection between the systems, not a depiction of an unchanged Cradle implementation controlling every aspect of a humanoid.

\subsection{Initial end-to-end realization}

Being-0~\parencite{being0} is an initial realization of the approach to and agentic end-to-end humanoid robot system. In it, a foundation model (GPT-4o) handles instruction understanding, task decomposition and reflection; a lighter fine-tuned VideoLLaMA2 vision-language model plays the role of connector and grounds the resulting intentions in egocentric RGB observations; and a library of executable skills realizes locomotion, manipulation and other actions. The original deployment uses learned manipulation policies, inverse kinematics, locomotion control and Python-based skill compositions. The connector and skills run on the robot.

\begin{figure*}[!t]
    \centering
    \includegraphics[width=.96\textwidth]{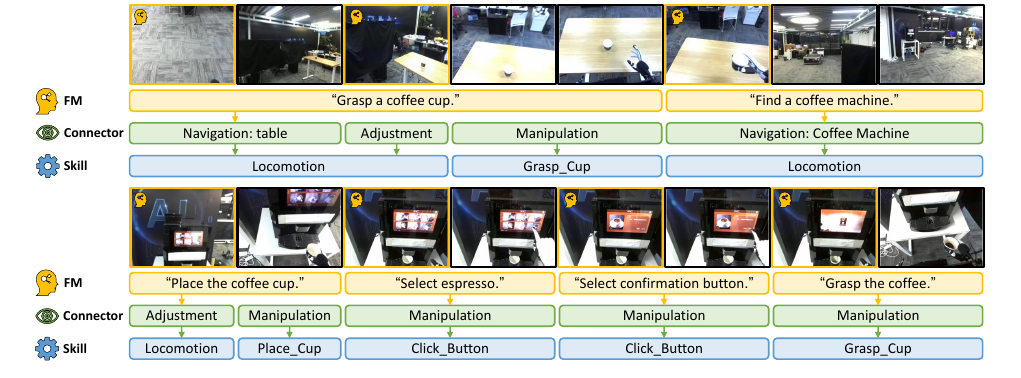}
    \caption{Coffee preparation in Being-0, reproduced from Fig.~2 of \textcite{being0}. The original yellow, green and blue annotations identify decisions by a foundation model, a connecting grounding model, and skills from its skill library.}
    \label{fig:being0_coffee_execution}
\end{figure*}

Figure~\ref{fig:being0_coffee_execution} illustrates a previously demonstrated coffee-preparation task. It exposes the sequence of high-level decisions and the key role grounding and the proper coordination of motor-skill invocations take within one physical task. Being-0's experiments on navigation-to-manipulation transitions and active camera control illustrate why the interaction between capabilities matters: a navigation policy may reach the right region but leave the robot in an unsuitable stance for manipulation. Likewise, a fixed camera orientation suitable for walking may not expose the information needed to grasp an object. Mobile ALOHA~\parencite{mobilealoha} offers a complementary example of learning coordinated mobile and bimanual manipulation from demonstrations on a wheeled platform.

\section{Agentic capability architecture}
\label{sec:architecture}

\begin{figure*}[!t]
    \centering
    \includegraphics[width=.96\textwidth]{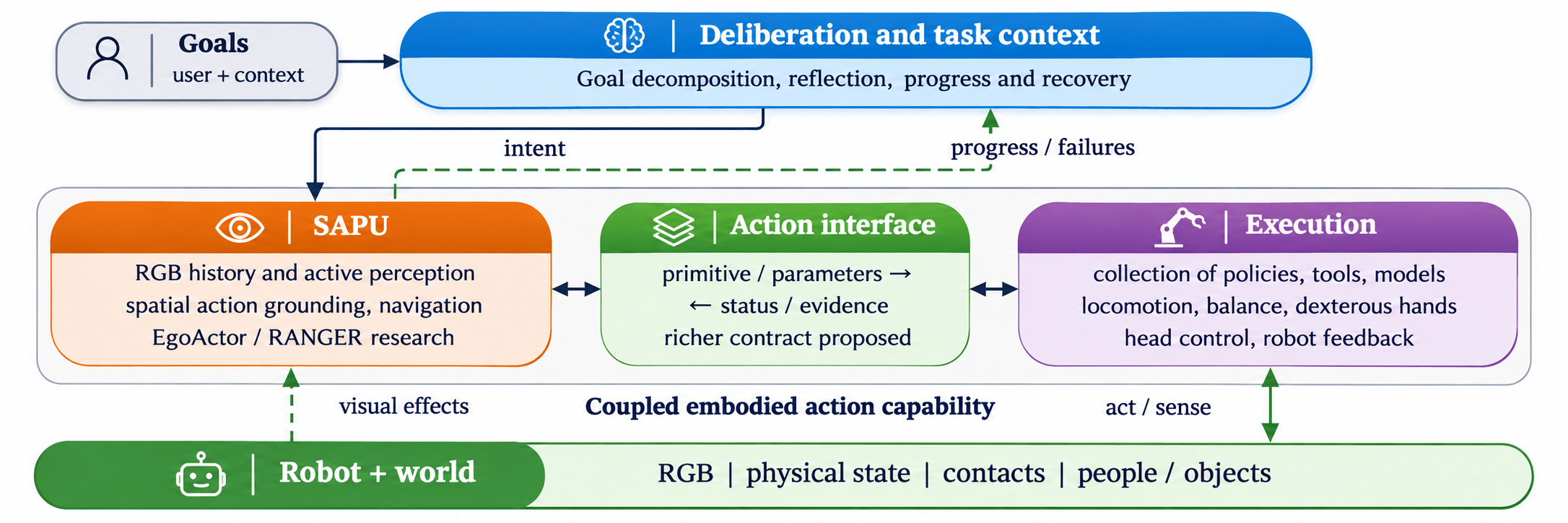}
    \caption{Current architectural organization. Deliberation works at task abstraction, while spatial action planning and execution are coupled aspects of embodied action. The action interface is bidirectional; existing primitives and policy calls motivate a richer contract.}
    \label{fig:current}
\end{figure*}

Different kinds of decisions are made at different abstractions and timescales. A robot must preserve the purpose of a task across many observations; choose spatially grounded actions that advance it; and realize these actions while remaining stable. We refer to these as Systems 2, 1 and 0, respectively (Figs.~\ref{fig:capabilities} and \ref{fig:current}). The terminology describes functional capabilities. It does not dictate the number of models, processes, layers, sub-systems, or sequential invocations. One model can implement several functions, or a capability can be distributed across models and tools. Explicit boundaries are useful while the required end-to-end behaviour is not yet well characterized. This architecture is both a concrete realization of our research approach and a means of discovering its additional requirements: the objective is a unified agentic system, not necessarily a single unified model.

\paragraph{Deliberation and task context (System 2)} A task-level agent processes and maintains instructions, relevant context, hypotheses about the world, task progress/milestones, and unresolved conditions. It plans and revises subgoals, decides when to ask for information, and evaluates evidence for task completion. It need not continuously emit motor commands. Its decisions may be informed by visual observations, reports from other modules or tool use, and retrieved experience. When a subgoal fails, it should distinguish a temporary execution problem from a reason to revise the goal decomposition. A plan is a revisable commitment, not an open-loop script. In \robonoid, a VLM as high-level planner and self-reflection instantiates this capability.

\paragraph{Spatial action planning (System 1)} The Spatial Action Planning Unit (SAPU) is a core component of \robonoid that grounds intent in egocentric observations, recent action history, task-relevant geometry and available execution capabilities. It may select navigation waypoints, change head orientation, request a manipulation, refine the target, or trigger a human interaction. This is more than converting text into the name of a skill: it includes understanding spatial relationships, estimating scale, coordinating transitions, deciding when additional evidence is needed and judging whether a short-horizon action advanced the task. In \robonoid, EgoActor~\parencite{egoactor} and RANGER~\parencite{ranger} instantiate complementary parts of this capability, with EgoActor covering the core responsibilities of the SAPU.

\paragraph{Physical realization (System 0)} The execution side comprises policies, tools, models, and controllers rather than necessarily one universal motor policy. Locomotion, balance, dexterous manipulation, posture, gaze, and platform-specific feedback can be served by distinct backends or by a shared whole-body model. The relevant robot state for a given skill may include proprioception, contact, or other safety-critical signals. Its responsibility is to realize feasible actions under physical constraints and report what is actually achieved. System 1 and System 0 are two sides of embodied action, not unrelated stages; a request that cannot be physically realized must change the grounding decision, and the resulting viewpoint and motion continuously change its inputs. In \robonoid this level is defined by a set of pre-trained manipulation models, an IK planner, a set of locomotion policies (decoupled upper/lower body or whole body), and tools for TTS, ASR, etc.

The natural runtime is a set of interacting loops. Execution can maintain balance while perception gathers new evidence; spatial grounding can adjust an approach while a task-level agent is evaluating longer-horizon progress. A new observation, infeasible target, timeout, human intervention or changed subgoal can trigger local revision or escalation. There is no requirement that every motor action be preceded by a new task plan. Feedback is not a separate final stage but a source of information at all abstraction levels.

Some actions can safely overlap (for example gaze and approach), while others require a verified precondition (for example entering a region after a door is sufficiently open). The executor must not rely on a slow remote model for stabilizing control. The system must reject stale or incompatible requests, preserve safety constraints, and terminate or degrade to a verified safe behaviour when feedback is unavailable. Actual scheduling and control frequencies depend on the deployed hardware and models; no universal timing claim is made here.

\subsection{High-level planning and self-reflection}
\label{sec:hlp}

The task-level planner is responsible for translating an abstract instruction into a feasible and revisable set of subgoals, rather than a fixed sequence of motor commands. It must understand intent and scope, reason over the environment and available capabilities, track progress, and decide when a subgoal should be revised or abandoned. For the running coffee-delivery example, failure to find the pantry, obstruction during approach, pressing the wrong control, dropping the cup, or failure to locate the recipient's desk may each require a different response. Merely generating a more plausible initial plan is insufficient if the planner cannot diagnose a stalled or invalidated subgoal. Grounding planning in multimodal observations and feedback builds on earlier embodied reasoning and affordance-based approaches~\parencite{palme,saycan,inner}.

\begin{figure}[!t]
    \centering
    \includegraphics[width=1.0\columnwidth]{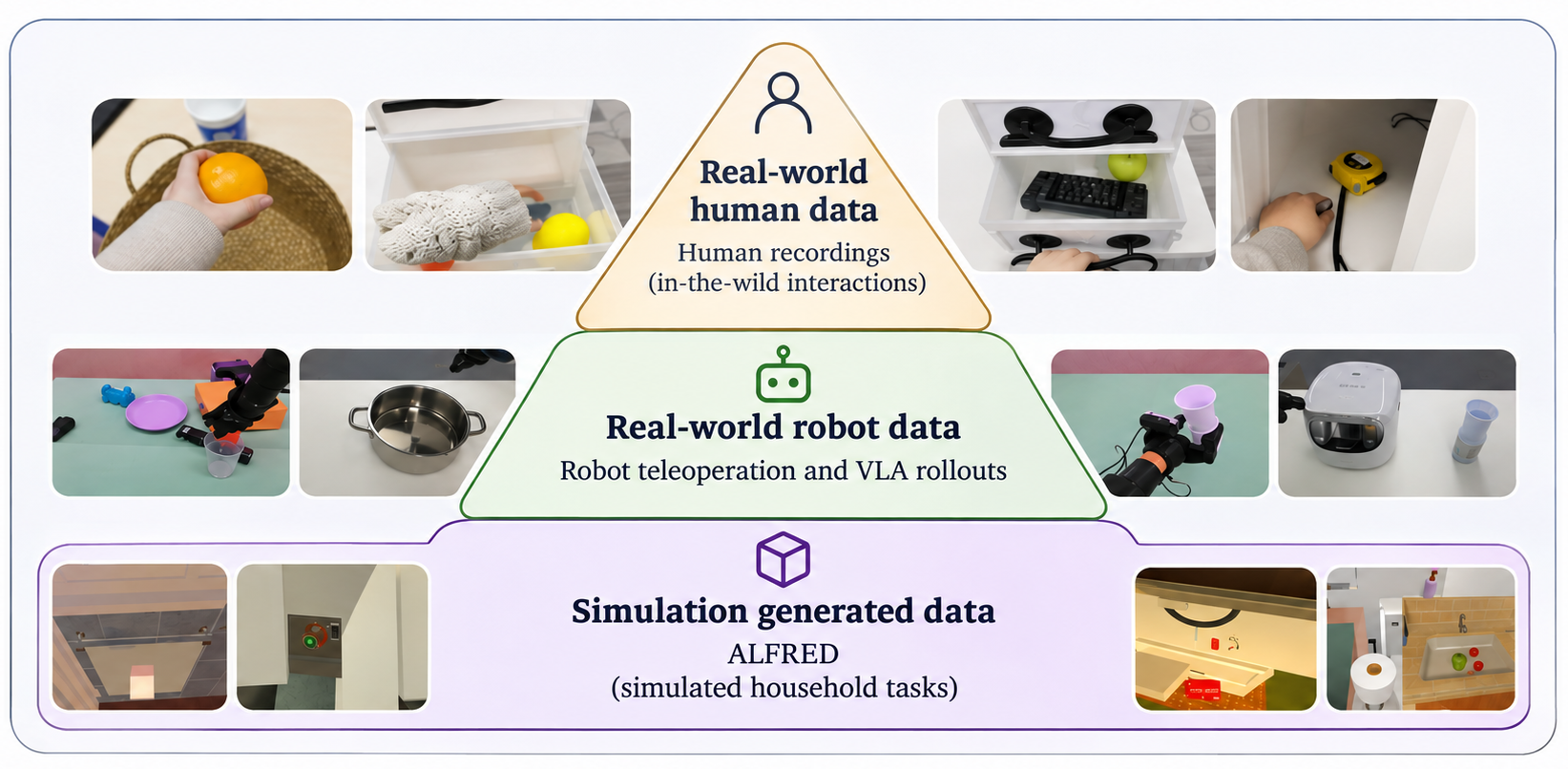}
    \caption{Examples of the heterogeneous sources used in HLP reflection-data construction: human recordings, robot teleoperation and VLA rollouts, and ALFRED-generated simulation trajectories. The counts describe the development dataset.}
    \label{fig:hlpdata}
\end{figure}

The current high-level planner (HLP) uses a Qwen3-VL-8B model with LoRA-based supervised fine-tuning. Its training recipe combines adapted existing datasets, synthetic examples, and collected or edited real executions. An initial training stage targets planning and task decomposition using approximately 900,000 samples, followed by a second stage targeting self-reflection using approximately 16,000 samples. The latter draws on human recordings, robot trajectories and simulated executions (Fig.~\ref{fig:hlpdata}). These approximate counts describe the current development recipe rather than a finalized training or evaluation specification.

Reflection supervision covers cognition-level failures (such as infeasible plans, ambiguity and hallucination), execution-level failures (such as incorrect pose, incomplete actions and loss of balance), and observation-level failures (such as occluded objects and unexpected obstacles). The intended behaviour is to assess whether the plan remains reasonable, summarize progress and evidence for successful or failed steps, and adapt or revise the plan when appropriate. The evaluation of reflection must ultimately consider its effect on subsequent decisions and task completion, not only the quality of its textual explanations. Initial planning experiments have been conducted, but the integrated effect of reflection is not quantified here. Proprio-MLLM~\parencite{proprio} provides a complementary investigation of embodiment-aware planning, where access to robot-state information helps reason about feasibility and contingencies.

Grounding also extends beyond the current egocentric view. In the current RoboNoid implementation, the high-level planner receives a JSON representation of the connectivity of the learned indoor layout, including objects from the semantic graph grouped by region. This provides explicit global context for resolving spatial references and selecting task-relevant destinations. Visual maps can complement this representation with spatial geometry, as discussed in Sec.~\ref{sec:rgbmap}.

\begin{figure}[!t]
    \centering
    \includegraphics[width=\columnwidth]{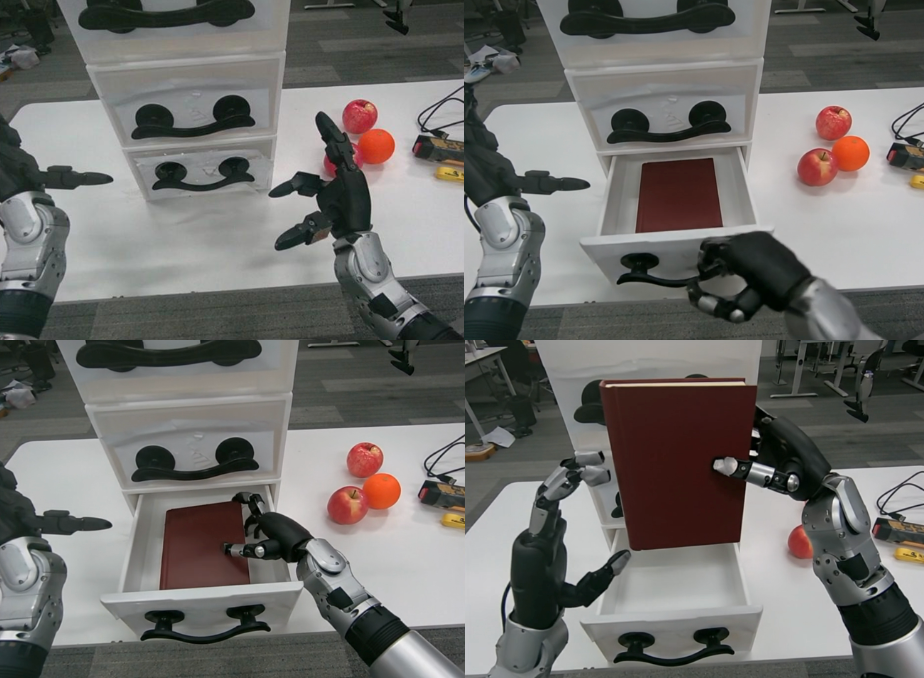}
    \caption{Egocentric observations illustrating drawer opening and retrieval of a book. The frames show the manipulation sequence for a drawer containing the target; the preceding unsuccessful search of another drawer is part of the illustrative HLP scenario, not depicted here.}
    \label{fig:drawer_search}
\end{figure}

\paragraph{Running example: searching for an object in drawers} Consider the instruction to retrieve a book from a set of drawers. The HLP decomposes the task into identifying a candidate drawer, opening it, inspecting its contents and, if the book is present, retrieving it. If the first drawer does not reveal the book, successful opening is not sufficient to mark the retrieval subgoal as complete. The planner should record the negative observation, reconsider the remaining candidate locations, and request inspection of another drawer rather than repeat the same search. Once the book is found, SAPU grounds the grasp and the execution capability realizes it; the resulting observation provides evidence of retrieval. This example illustrates how information-gathering actions change the task plan and how working memory prevents redundant search. Figure~\ref{fig:drawer_search} shows the opening and retrieval portion; the unsuccessful first search is illustrative and is not shown in the supplied frames.

\subsection{Spatial action planning and active perception}
\label{sec:sapu}

\begin{figure*}[!t]
    \centering
    \includegraphics[width=.96\textwidth]{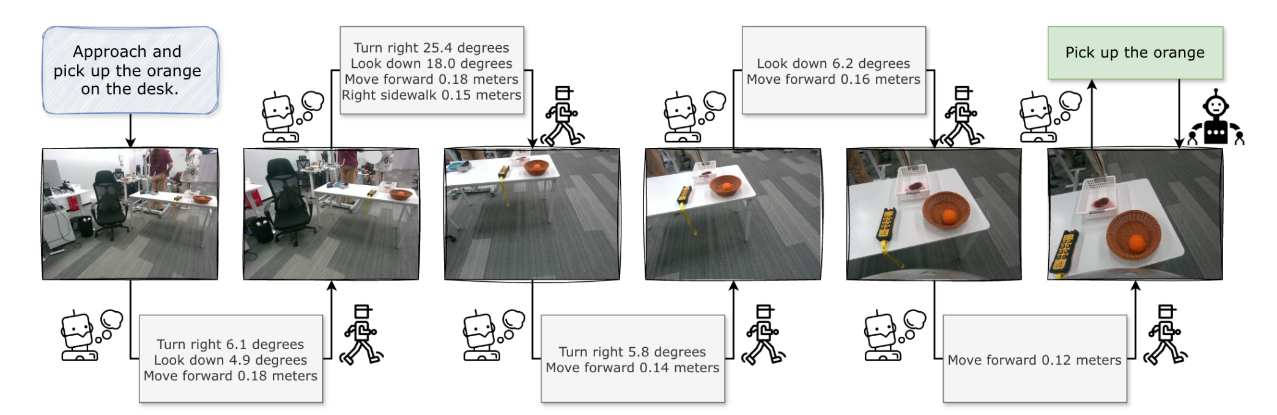}
    \caption{Example of spatial action planning from an egocentric observation sequence, at the SAPU level. A request to approach and pick up an orange is progressively grounded in movement, viewpoint, and manipulation commands. The figure illustrates coordination and command generation, including adjustments and retries. Reproduced from~\textcite{egoactor}.}
    \label{fig:sapu_example}
\end{figure*}

EgoActor~\parencite{egoactor} realizes a central part of SAPU: translating a task-level instruction into spatially grounded decisions from a history of egocentric RGB observations. The model is based on Qwen3-VL (4B or 8B) and can produce parameterized movement and viewpoint commands, manipulation requests, and actions related to human interaction. This permits the agent to coordinate relatively coarse locomotion, progressively finer positioning, head orientation and skill invocation without requiring a new high-level plan at every step. Figure~\ref{fig:sapu_example} shows an example of such a sequence.

The use of an active camera is important because the information needed for the next action depends on its purpose. When approaching an object, the robot may change its head orientation to refine the estimated target and the relationship between the object and its own body. When identifying a person, it may look up to inspect clothing or other attributes. The relevant distinction is between choosing a useful observation and demanding that every observation already contain all necessary information. Language-conditioned navigation and spatial representations offer complementary ways to ground such decisions~\parencite{navila,vlmaps}.

\begin{figure*}[!t]
    \centering
    \includegraphics[width=.90\textwidth]{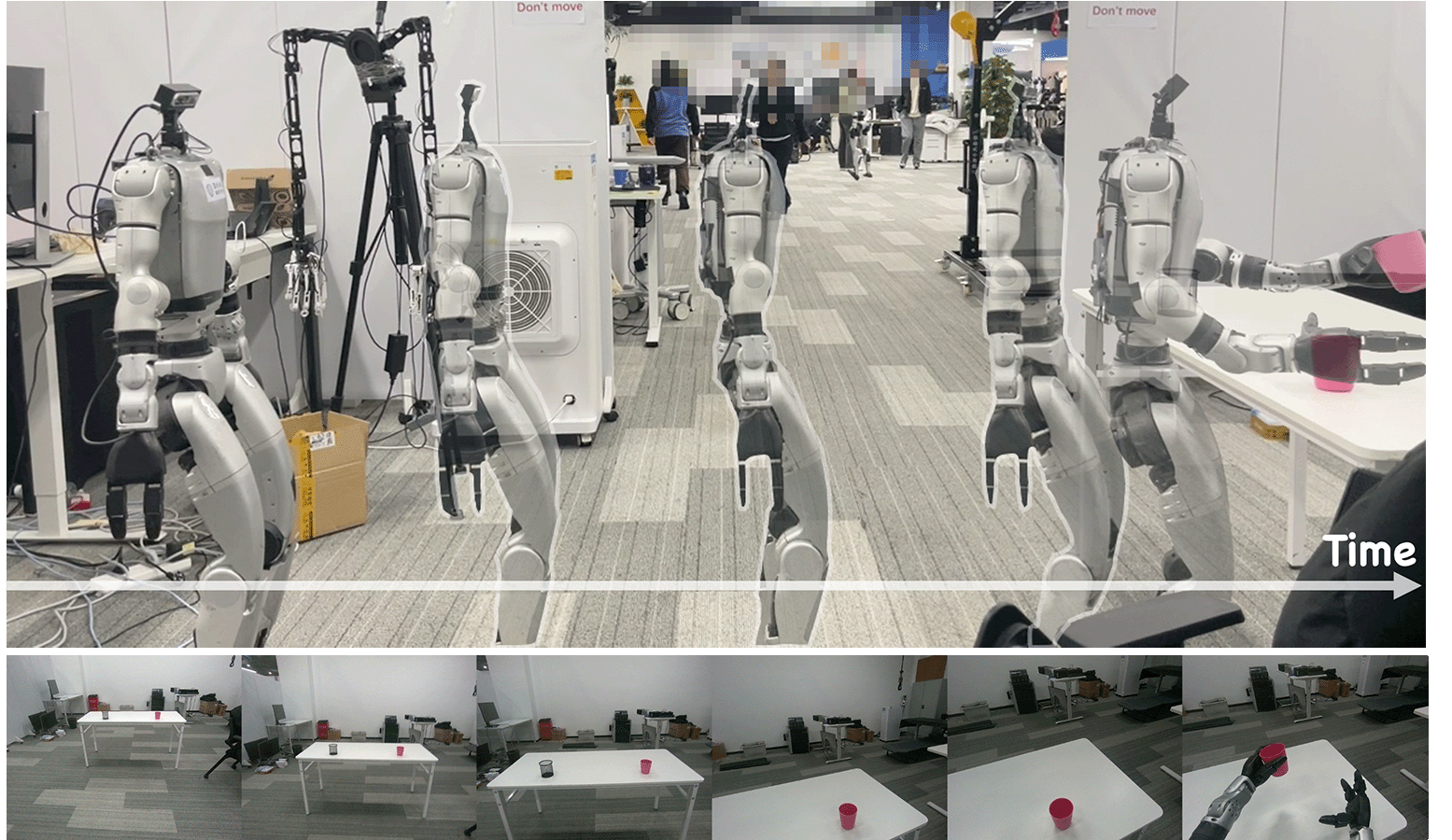}
    \caption{Real-world mobile manipulation with EgoActor: approaching and grasping a previously unseen pink cup. The upper sequence shows the physical humanoid; the lower sequence shows its egocentric observations. Reproduced from Fig.~9 of \textcite{egoactor}.}
    \label{fig:realrobot}
\end{figure*}

\subsection{Physical realization and the skill library}
\label{sec:skills}

The published EgoActor deployment provides a concrete reference implementation using a Unitree G1 with Dex3-1 hands, a two-degree-of-freedom head and a RealSense D455 operated in RGB-only mode. The spatial decision model issues requests through a dispatcher to a walking policy, head controller and fine-tuned GR00T N1.5 manipulation policies. Locomotion retains joint-state and inertial feedback. The lower-level policies run at a higher frequency than the spatial decision model~\parencite{egoactor}.

The current library includes active head movement, parameterized locomotion, manipulation skills (such as grasping, placing and opening or closing drawers), inverse-kinematics routines, prerecorded gestures and Python-code compositions. Some behaviours are realized by task-specific learned policies; others are implemented using existing controllers and tools. A library of this kind allows new capabilities to be added and tested without requiring the entire agent to be retrained. Programmatic skill composition~\parencite{cap} and generalist VLA policies~\parencite{openvla,grootn1} provide different mechanisms for implementing such reusable behaviours.

Existing systems can invoke a fixed skill with language or spatial parameters. This is a useful starting point, but a more capable action interface should convey intent, targets or regions, constraints, optional behaviour priors, expected effects and termination conditions. Equally important is the reverse interface: feasibility, actual execution status, estimated achieved state, violations and evidence. We call the proposed representation an \emph{embodied action contract}. It is a typed exchange, not a separate inference stage or a demand for a particular control algorithm. The grounding side expresses what is needed; the execution side can accept, reject, refine within allowed bounds, or report a failure.

For a reach toward a cup behind the edge of a table, a grounded request might specify a target region and a conservative exclusion region derived from RGB rather than a fully generated trajectory. The backend can solve the motion while preserving stability, or report that the target is unreachable from its stance. The agent may change its stance or viewpoint and update the request. A controller's successful termination is not evidence that the object was grasped; verification may require a visual observation and, where available, hand-state evidence. This separation allows heterogeneous control approaches, including structured primitives, geometric goals and learned latent actions, to coexist and potentially converge.

\section{Examples of coordinated behaviour}
\label{sec:examples}

\begin{figure*}[!t]
    \centering
    \includegraphics[width=.80\textwidth]{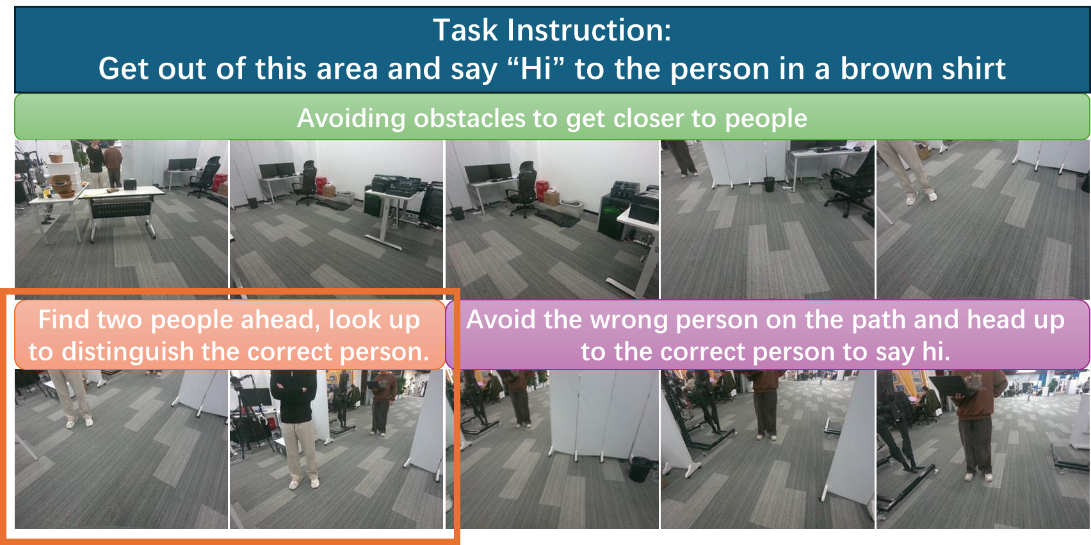}
    \caption{Egocentric observations from a human-interaction example. The task calls for approaching the person in a brown shirt; active perception helps disambiguate the target and guide the subsequent approach. Reproduced from~\textcite{egoactor}.}
    \label{fig:personinteraction}
\end{figure*}

Representative real-world scenarios in our development demonstrations include entering a meeting room, navigating around obstacles and people, identifying a person based on a description, and approaching a desk for manipulation. Figure~\ref{fig:personinteraction} illustrates a human-interaction task in which the agent changes its viewpoint to resolve which person the user intends it to approach. These examples are valuable for identifying dependencies between spatial reasoning, active perception and physical execution. They are illustrative demonstrations, not a standardized evaluation or an estimate of success rate across environments.

In mobile manipulation, correctly selecting and triggering a skill is not necessarily equivalent to accomplishing the intended physical effect. The EgoActor study distinguishes spatial grounding from physical manipulation outcomes~\parencite{egoactor}. This motivates retaining evidence of what actually occurred and allowing the agent to revise its subsequent actions accordingly.

\section{RGB-based layout understanding and map learning}
\label{sec:rgbmap}

Layout understanding is necessary when a task extends beyond the current field of view. The agent must retain where it has been, which regions have been explored, where relevant entities were seen, and how to reach them. We distinguish this environmental knowledge from both the transient egocentric action context and an exhaustive reconstruction of the scene. RANGER provides a concrete implementation of incrementally acquiring actionable spatial knowledge from monocular RGB \parencite{ranger}. Its goal is open-vocabulary object navigation, not dense reconstruction as an end in itself. Related work on visual-language maps and open-vocabulary scene graphs demonstrates the utility of spatially organized semantic features for language-conditioned navigation and planning~\parencite{vlmaps,conceptgraphs}.

\begin{figure*}[!t]
    \centering
    \includegraphics[width=.96\textwidth]{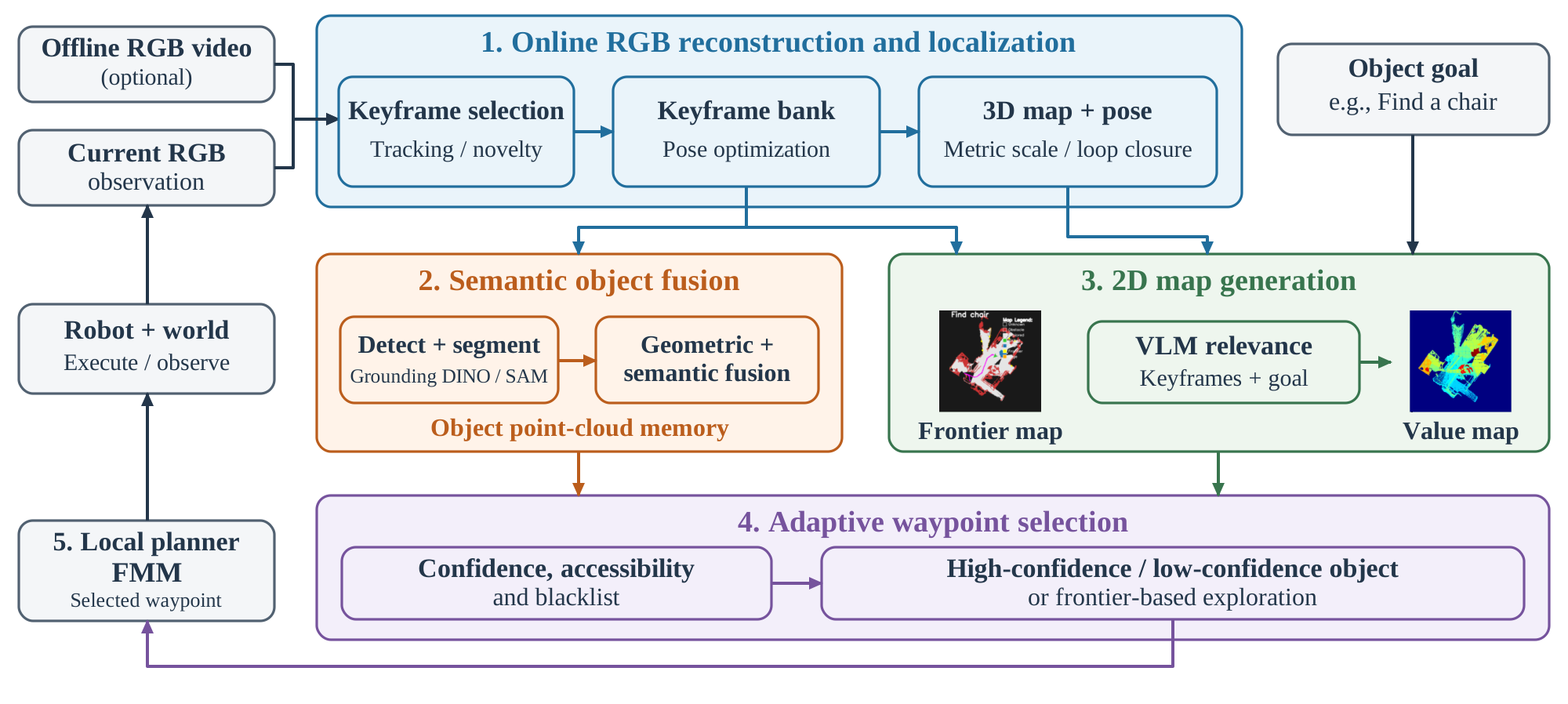}
    \caption{RANGER mapping and navigation pipeline~\parencite{ranger}. A keyframe bank links RGB reconstruction and localization, semantic fusion, frontier and value maps, and adaptive waypoint selection. Offline traversal video is optional.}
    \label{fig:rangermap}
\end{figure*}

As shown in Fig.~\ref{fig:rangermap}, RANGER uses MASt3R-SLAM to select keyframes based on novelty and tracking confidence. Dense pointmaps inferred from RGB are matched to the current keyframe; relative pose optimization and loop closure maintain a pose graph and refine localization as observations arrive. Monocular geometry has a scale ambiguity: RANGER uses known camera height and a fitted ground plane to recover metric scale. Thus, the method does not require a depth sensor or externally supplied accurate poses, but does use inferred geometry and a physical scale prior.

For semantic mapping, Grounding DINO and MobileSAM detect and segment objects within keyframes. Reliable reconstructed points are selected using confidence masks; CLIP features describe the corresponding image crops. Cross-frame associations combine point-cloud overlap and feature similarity to decide whether to initialize a new object or merge observations into an existing one. Projection of the reconstructed point cloud produces a two-dimensional obstacle and explored-area map, while relevance scores computed against the target description produce a value map. Candidate waypoints include object proposals and exploration frontiers; semantic confidence, accessibility and a blacklist of previously unreachable targets guide selection. A Fast Marching Method planner generates local paths, with the map and target updated as execution unfolds.

An optional traversal video can initialize the keyframe bank before the robot starts. RANGER investigates how previously recorded visual experience can support subsequent navigation without requiring that the recording be synchronized with the current robot's actions. The published contextual-video experiments use overlapping observations to support relocalization~\parencite{ranger}. This is an example of decoupling data acquisition from the embodiment and task for which it will later be used.

\begin{figure*}[!t]

\centering
\begin{minipage}[t]{.32\textwidth}
\centering
\includegraphics[width=\linewidth,height=1.55in,keepaspectratio]{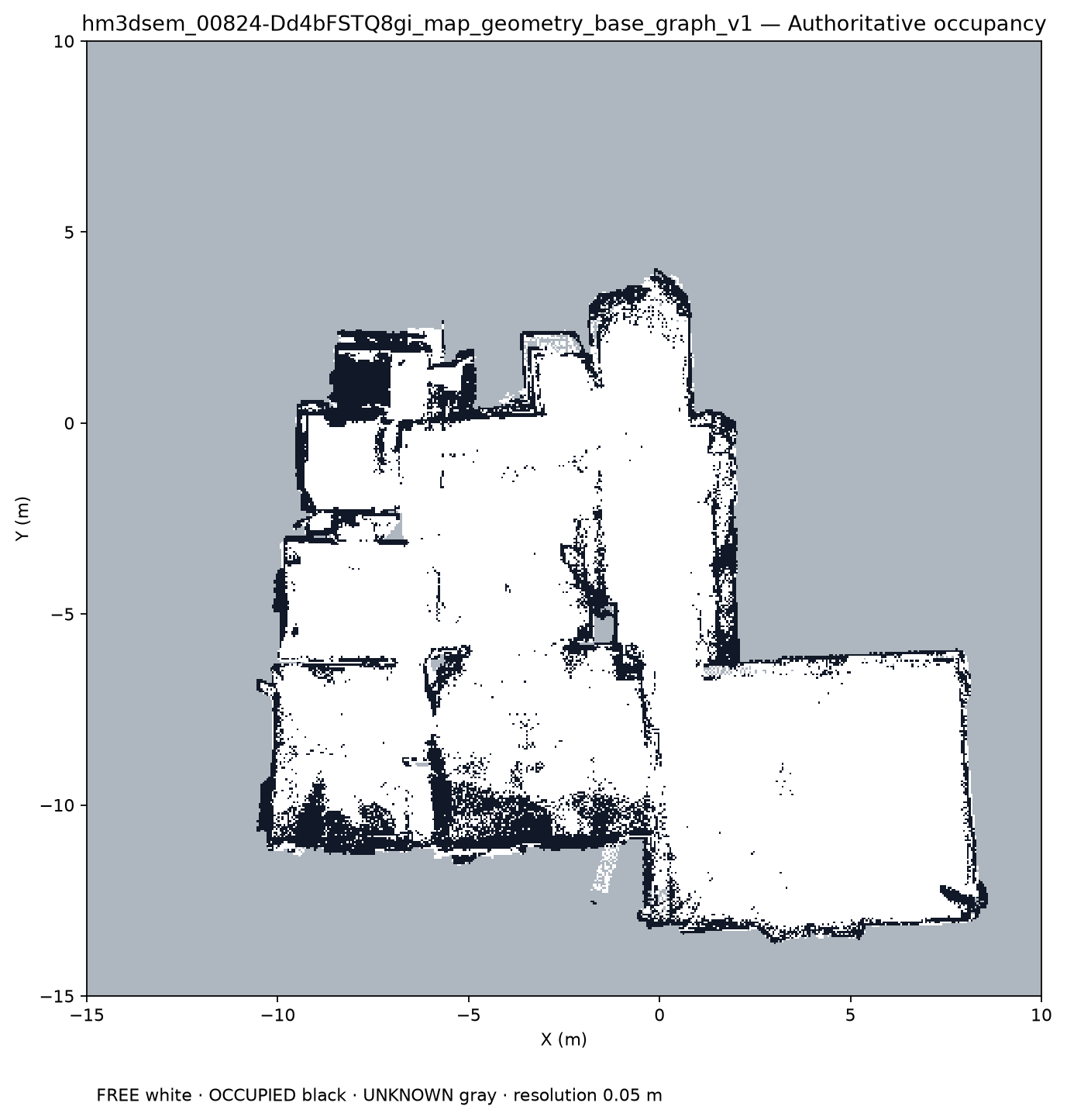}\\
{\small (a) Occupancy projection}
\end{minipage}\hfill
\begin{minipage}[t]{.32\textwidth}
\centering
\includegraphics[width=\linewidth,height=1.55in,keepaspectratio]{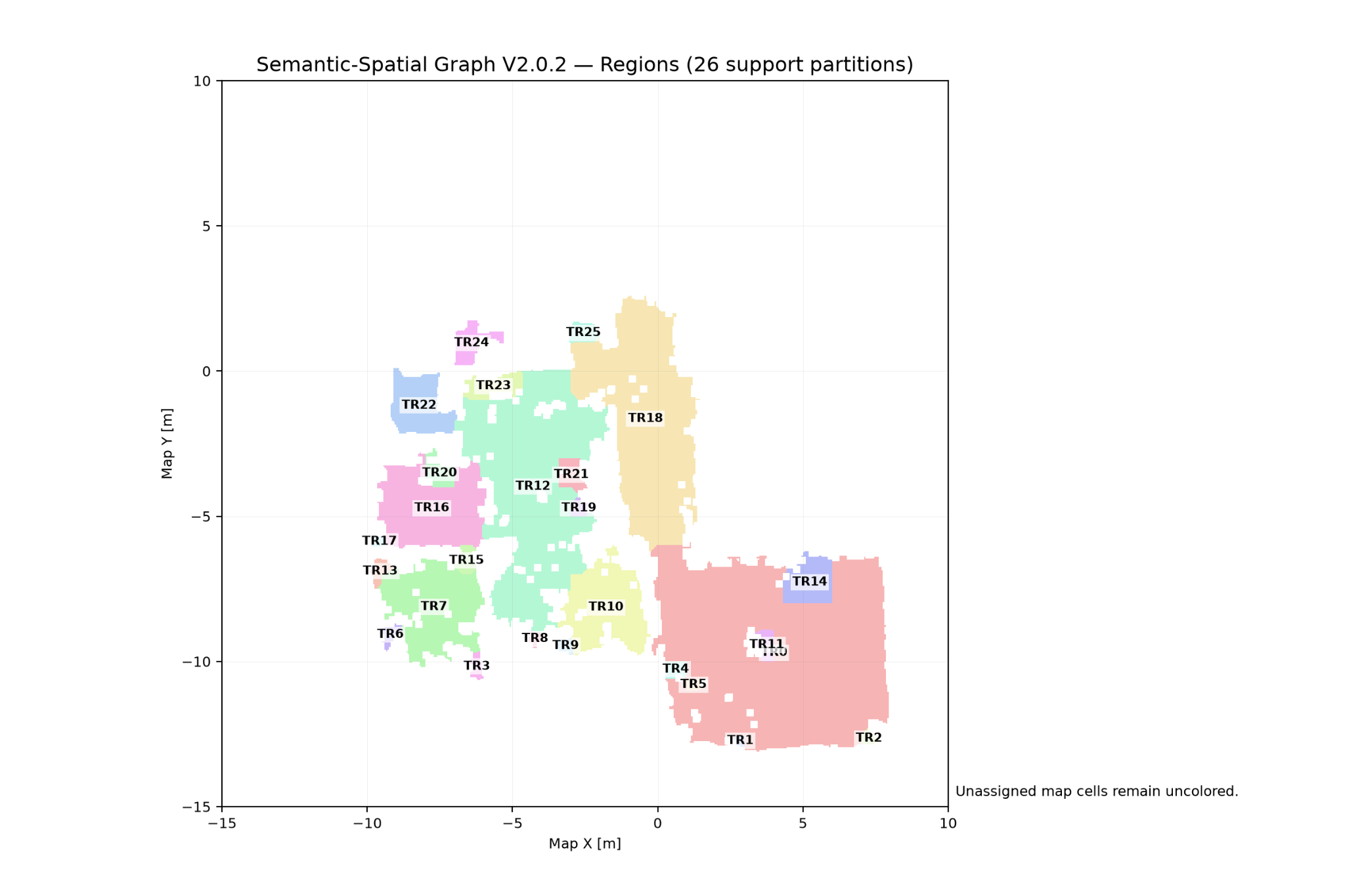}\\
{\small (b) Support-region partitions}
\end{minipage}\hfill
\begin{minipage}[t]{.32\textwidth}
\centering
\includegraphics[width=\linewidth,height=1.55in,keepaspectratio]{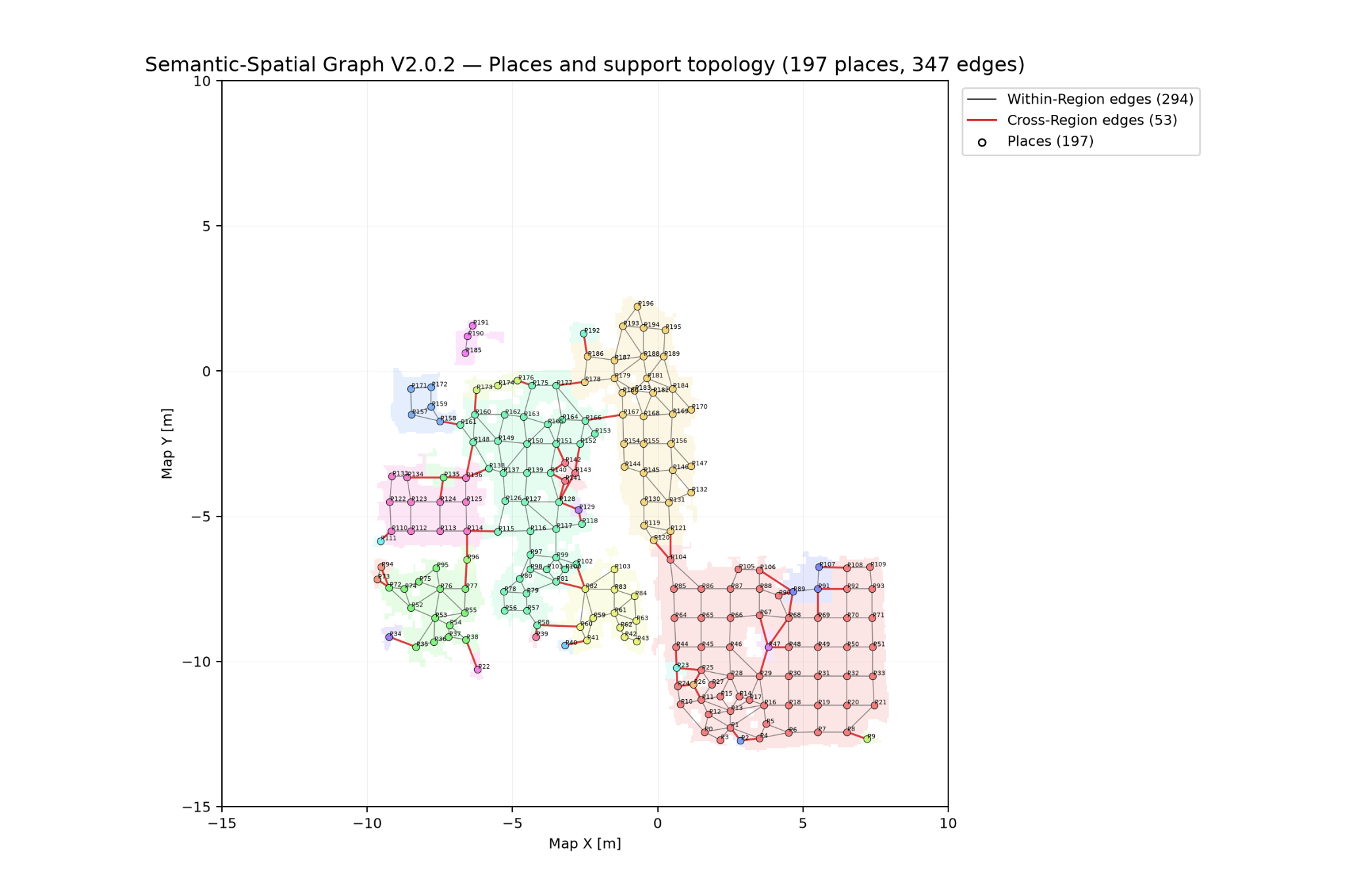}\\
{\small (c) Place connectivity}
\end{minipage}
\caption{Example spatial representations from the current RANGER development, using an HM3D scene: occupancy projection, support-region partitions and place connectivity. Semantic object locations can also be projected in the same coordinate frame.}
\label{fig:rangerlayout}

\end{figure*}

In the current development of RANGER for RoboNoid, exploration and mapping can also be invoked as skills for acquiring environmental knowledge. Spatial memory extends beyond the original pose graph and semantic point clouds to projected occupancy maps, support-region partitions, and place connectivity. Object positions from the semantic graph can be projected into the same coordinate frame. Figure~\ref{fig:rangerlayout} illustrates these representations, which provide geometric, topological and semantic context at different levels of abstraction. The learned layout connectivity and objects associated with each region are supplied to the high-level planner in JSON form. A connectivity edge does not guarantee current traversability; local observations remain necessary to adapt to changes in the environment.

Acquiring a map and understanding a map are distinct capabilities. MURIS~\parencite{muris} investigates targeted supervision for ego-to-map localization, language-conditioned target grounding and waypoint planning on top-down RGB maps. It motivates further training of the high-level planner to interpret visual maps, whether supplied independently or produced by RANGER, as a complement to the structured layout representation. MURIS does not establish transfer to occupancy-only projections or incomplete reconstructions; integration of these visual inputs with the planner is an ongoing research question.

\section{Scaling by reducing constraints}
\label{sec:scaling}

Egocentric RGB observations are comparatively inexpensive to acquire from human activity, autonomous robot operation, teleoperation, simulated scenes, and existing video collections. They do not require a common robot morphology or synchronized force and joint measurements. Nor special sensors or multiple data feeds. This makes them a useful common currency for learning the aspects of action that can be inferred visually: relationships, affordances, local geometry, temporal changes, and task-relevant effects. R3M~\parencite{r3m} illustrates the reuse of human-video representations for robot manipulation, while Open X-Embodiment~\parencite{openx} demonstrates the complementary value of pooling robot-specific experience across platforms.

The important qualification here is that common observations do not imply common labels. A video of a human using an appliance may support learning what the controls mean, but not the torque necessary to operate them with a robot hand. Camera motion, occlusion, embodiment mismatch and contact ambiguity can make several interpretations consistent with the same images. Generated videos may offer useful variation but can be physically inconsistent. Thus we propose to separate the information needed for decisions from the information required to execute them, rather than insist on a uniform data format for all supervision.

A useful acquisition strategy begins with widely available supervision, then adds more expensive in-robot evidence where behaviour exposes a need. For example, learning to recognize whether a mug is upright or inverted may not require motor data; estimating a viable grasp for a particular non-standard hand does. If the hand reaches the object but fails to secure it, a short instrumented correction may be more informative than another complete demonstration of the task. Neither the data ratio nor the benefit of that strategy is assumed in advance; it should be evaluated against alternatives at matched budgets.

\begin{figure}[!t]
    \centering
    \includegraphics[width=\columnwidth]{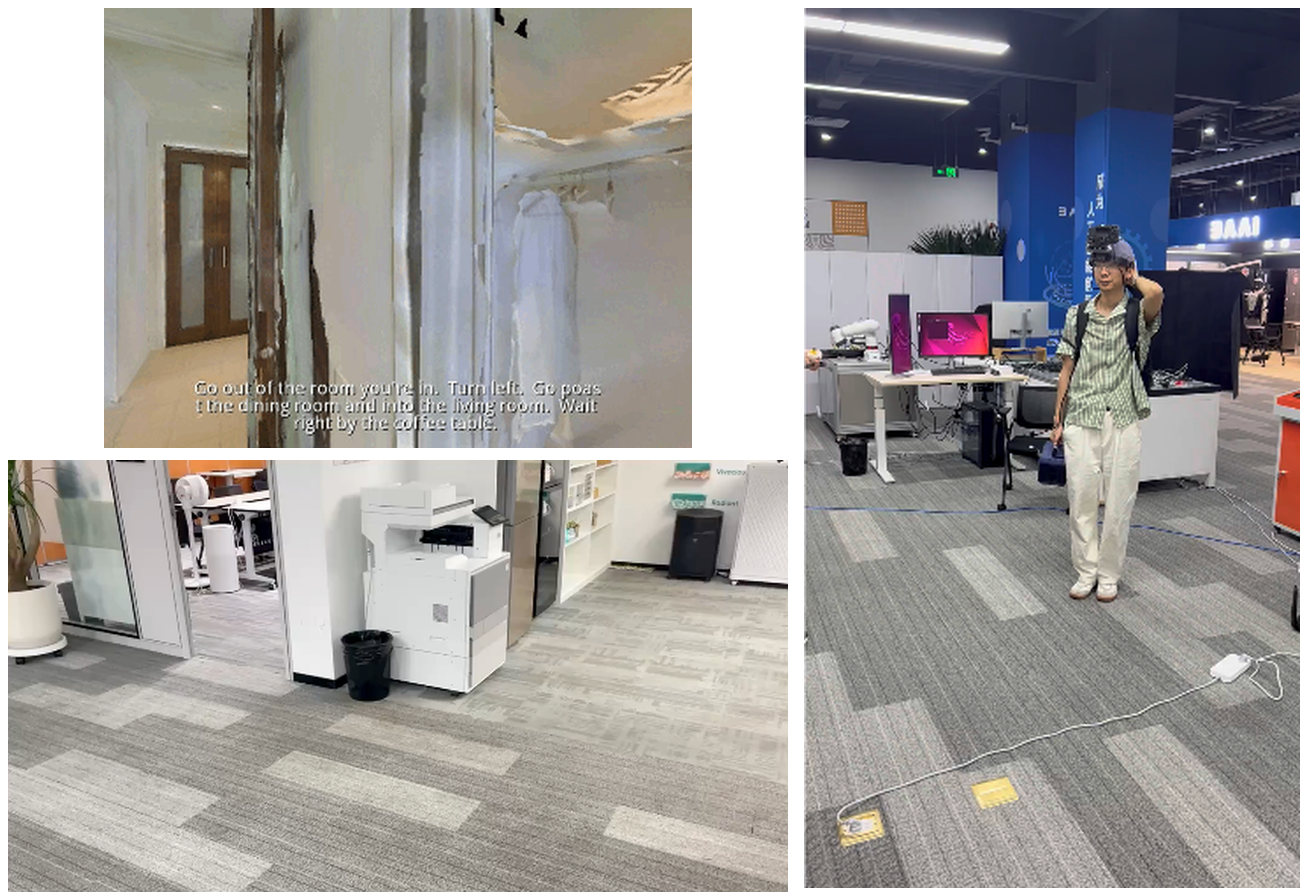}
    \caption{Examples of RGB data acquisition from simulation,
    egocentric recordings of real environments, and human-carried
    cameras. Such sources provide different kinds of supervision
    and do not all require robot-specific action labels.}
    \label{fig:rgb_sources}
\end{figure}

EgoActor provides a concrete example of mixed supervision: egocentric videos, virtual navigation and spatial-action trajectories, spatial reasoning, visual question answering, task planning and a smaller set of situated robot executions~\parencite{egoactor}. Camera motion can be estimated offline to derive movement supervision from video. This preprocessing is distinct from requiring a depth sensor or accurate poses at deployment. The resulting model coordinates spatial actions; robot-specific manipulation policies supply physical execution capabilities. The mixture illustrates the acquisition strategy, not a measured scaling law.

\subsection{Adaptation instead of unnecessary precision}

A sufficiently accurate perception system is not necessarily one that reconstructs every surface to high accuracy. It is one that supplies enough information for the next safe and useful decision. A coarse estimate of a work surface may suffice to select a view or approach region; contact-rich manipulation may demand a different sensing and control regime. The agent should be able to obtain additional observations when the expected uncertainty matters, and to avoid committing to an action when it cannot establish feasibility.

Errors can occur in both observation and execution. For instance, walking changes the camera viewpoint, the inferred position of an object may drift, and an achieved hand pose may differ from the requested pose. Re-observing the actual world, updating the target and correcting the action can be preferable to requiring a globally precise map or attempting to predict the entire motion in advance. This does not relax safety limits: balance, contact, collision, and hardware constraints must remain enforced by appropriate mechanisms. It relaxes assumptions about the accuracy of task-level predictions.

We distinguish three quantities: the breadth of useful data, the range of tasks and environments in which capabilities are reusable, and the cost of incorporating additional capabilities. A system may scale in parameter count while becoming more expensive to adapt to a new robotic hand or environment. Conversely, a small module may substantially improve many tasks if it resolves a frequent integration failure. A systematic study should therefore report acquisition cost, deployment cost, and end-to-end gains alongside conventional model scaling variables\footnote{We do not claim a quantitative scaling law in this report.}.

\subsection{Extending supervision and execution capabilities}

Generated visual experience offers a complementary avenue for expanding behavioural coverage. ExoActor~\parencite{exoactor} investigates third-person video generation conditioned on a task and a scene observation, extracting motion and hand-state references and realizing them through SONIC. It illustrates how imagined behaviours may be translated into physical motion, while leaving open important questions about geometric accuracy, contact and physical feasibility. Beyond generating nominal demonstrations, an ExoActor-like pipeline could also be used to propose counterfactual executions and plausible failure cases (e.g., a missed grasp, a changed object position or an unsuccessful approach) for reflection and recovery training. This is a possible extension rather than a demonstrated ExoActor capability: synthetic outcomes would need physical and semantic validation before being treated as reliable corrective supervision. Related work on video-based policies and interactive generative simulators provides complementary motivation~\parencite{universalvideo,unisim}. ViLoMan~\parencite{viloman} instead constructs whole-body loco-manipulation trajectories from partial human interaction demonstrations and generated approach motions, followed by physics-aware refinement and teacher--student training. Its deployed door-closing policy uses depth and proprioception. X-DiffVLA~\parencite{xdiffvla} explores sharing action-generation capabilities across embodiments. These are complementary investigations into the data and execution problem, not backends assumed to be fully integrated into RoboNoid.

\section{Shared state, evidence and memory}
\label{sec:memory}

\begin{figure*}[!t]
    \centering
    \includegraphics[width=.90\textwidth]{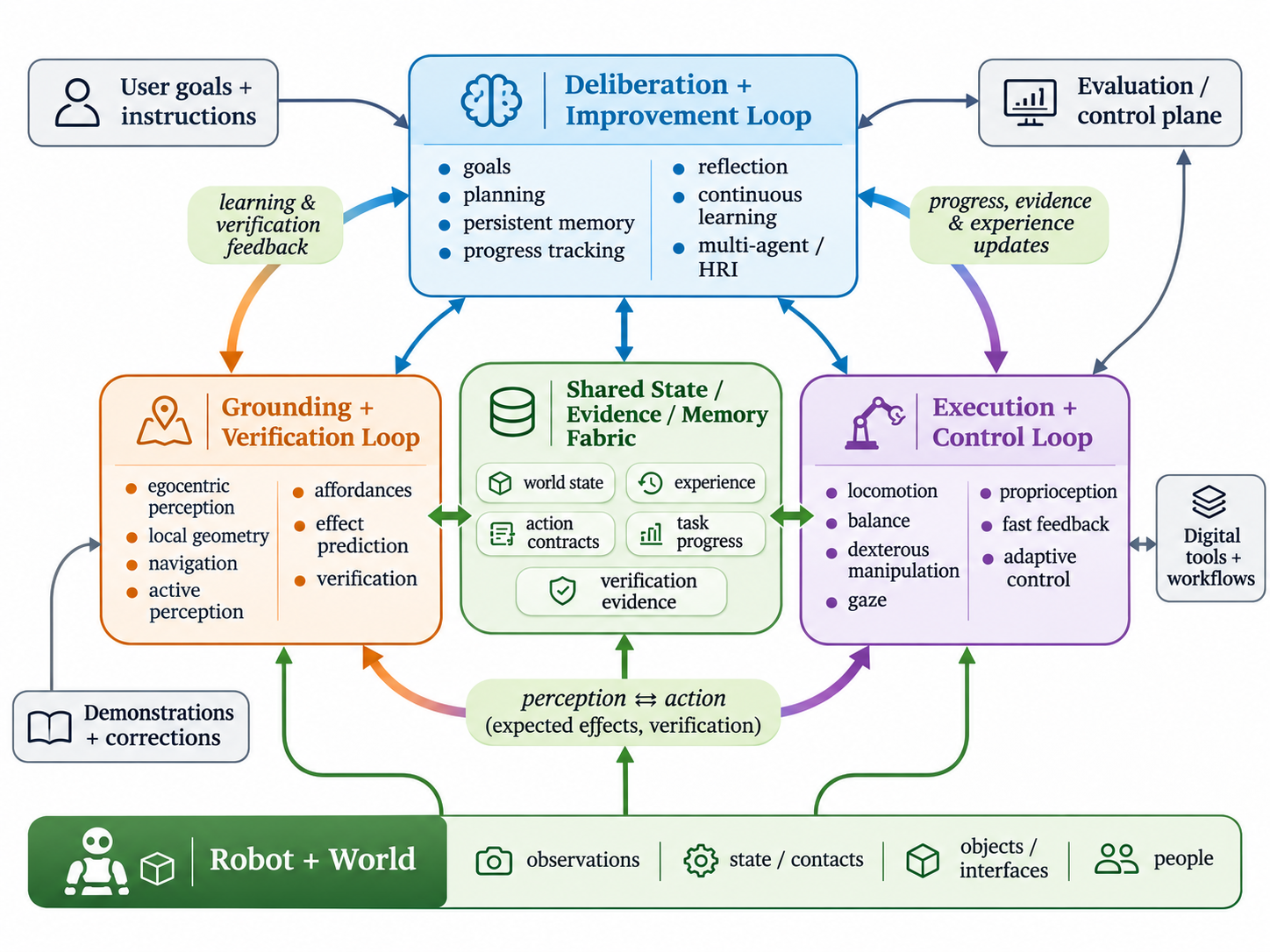}
    \caption{\robonoid's interacting capability loops and their shared state, evidence, and memory fabric. The illustration expresses functional communication and improvement paths rather than a fixed hierarchy or an inventory of fully deployed services.}
    \label{fig:memoryarchitecture}
\end{figure*}

Memory is not an optional context buffer attached to deliberation. Partial observability means that grounding and execution also depend on evidence that may no longer be visible, such as where the cup was left or whether a previous command changed the coffee machine's state. Our embodied memory survey distinguishes six requirements: hidden world state, action-induced state change, task state, delayed-relevance evidence, active belief maintenance and cross-episode experience \parencite{embmem}. These requirements are distinct from the mechanisms used to address them. A keyframe store, symbolic record, spatial graph, recurrent representation or procedural library can serve different requirements in combination. The architectural objective in Fig.~\ref{fig:memoryarchitecture} is shared access to decision-relevant state and evidence, not a single mandatory physical database or a requirement that all loops synchronously read and write the same representation.

\subsection{Working state and spatial memory}

We distinguish current working state from the historical evidence that supports it. Working state includes the active instruction and subgoals, their completion conditions, current environment and embodiment hypotheses, active action requests, constraints and pending verification. A report of a completed controller action is not necessarily evidence of the desired task effect. Grounding and execution supply observations, action status and outcomes; deliberation uses these to update progress, diagnose discrepancies and decide whether to continue, ask, recover or replan. Other capabilities should be able to retrieve the evidence behind a state assertion when the relevant detail is not in their immediate context.

The RANGER keyframe bank (Sec.~\ref{sec:rgbmap}) provides a concrete form of spatial memory: visual evidence and inferred geometry are retained across viewpoints, with object associations, region connectivity and projected maps supporting navigation and task-level reasoning. As the environment changes or localization is revised, earlier associations may become invalid. The shared state should distinguish an observation from an inferred location, and retain the evidence necessary to reconsider that inference. Graph-based semantic maps such as ConceptGraphs~\parencite{conceptgraphs} provide a complementary example of compact representations linking entities and spatial relations.

\subsection{Episodic and procedural memory}

Episodic memory preserves histories of observations, goals, decisions, actions and outcomes, including unsuccessful attempts and corrections. MART~\parencite{mart} studies learning to retrieve trajectories based on their downstream usefulness and abstracting them into salient milestones and actions. This provides a precedent for selecting relevant experience rather than simply retrieving the most visually or semantically similar episode. Reflexion and ExpeL~\parencite{reflexion,expel} illustrate how episode-level feedback and extracted lessons can guide later decisions without updating the underlying language model.

Procedural memory stores reusable knowledge of how to accomplish a task or recover from a failure, distinct from recalling a specific episode. We distinguish a procedural skill (a strategy or parameterized sequence with applicability conditions and expected effects) from the executable motor policy or tool that realizes an action. PRACTICE~\parencite{practice} investigates a trainable skill learner that maintains a persistent procedural library using successful and failed trajectories. Its executor remains frozen while the library and the learner are progressively refined. This form of improvement is demonstrated in high-level embodied benchmarks; transferring it to physical humanoids requires grounding the procedural knowledge in the actual execution capabilities and validating its effects. Voyager~\parencite{voyager} and Agent Workflow Memory~\parencite{awm} demonstrate complementary forms of persistent executable skills and reusable routines in game and web environments, respectively.

\section{Situated experience and architectural evolution}
\label{sec:adaptation}

An agent that notices a failure but repeats the same action without changing its assumptions is not meaningfully adaptive. A practical loop starts with an expected effect, obtains evidence of the actual effect, updates the task state and chooses whether to continue, correct locally, request information or replan. Failures can originate in semantic intent, perceptual ambiguity, an invalid spatial target, controller limitations, timing or a changed world. The initial diagnosis is a hypothesis and may itself be wrong. The ability to acquire disambiguating observations and test alternative explanations is therefore important.

Consider a robot that has been asked to operate a coffee machine. The relevant state is not merely its hand position: the machine may be waiting for a cup, warming up, showing an error or have a different button layout. Repeating a successful button-press motion need not advance the task. SWITCH and DeepPHY provide complementary evidence that reasoning about effects, state transitions and recovery remains difficult \parencite{switch,deepphy}. An adaptive agent needs to interpret the outcome in context, including delayed effects and causal relations beyond immediate physics (e.g., human instructions and device conventions).

\begin{figure*}[!t]
    \centering
    \includegraphics[width=.90\textwidth]{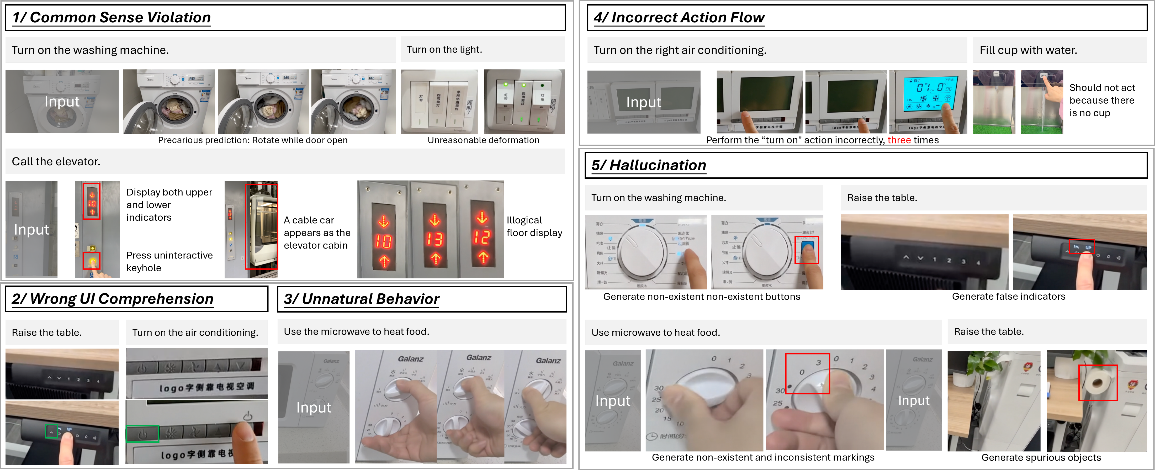}
    \caption{Examples of reasoning failures on tangible interfaces from SWITCH~\parencite{switch}: incorrect interface understanding, implausible behaviour, incorrect action flow and hallucination. These are diagnostic examples rather than robot execution failures.}
    \label{fig:switch_examples}
\end{figure*}

We view situated experience as more than a stream of images or an additional training corpus. The trajectory ties an instruction, observations, decisions, actions and their effects to the specific circumstances under which they occurred. It can reveal which capabilities are missing, which assumptions fail at their interfaces, which observations are sufficient, and which corrections are reusable. Successful and failed trajectories are valuable for different reasons. A failed grasp may identify a poor approach policy; a successful retry may reveal the relevant change in viewpoint or stance. Without outcome evidence, the learning system may reinforce a plausible but ineffective action. The use of previous attempts and feedback to refine subsequent behaviour is central to Reflexion, ExpeL and Voyager~\parencite{reflexion,expel,voyager}; RoboNoid extends the research question to physical actions, whose outcomes depend on the changing world and embodiment.

A useful experience representation records intended goals and effects, observations and timestamps, action requests and revisions, actual outcomes, intervention provenance, and uncertainty. It must distinguish observations from inferred state, and distinguish memory of an episode from a generalized rule. Episodic observations may be retrieved directly, distilled into procedural knowledge, or used to generate supervision for model updates. MART investigates interaction-grounded multimodal retrieval \parencite{mart}. PRACTICE learns a dedicated skill-library update policy from successful and failed task trajectories while keeping the executor frozen \parencite{practice}. It demonstrates a particular form of co-evolution (the learner and library), not continuous adaptation of a full humanoid controller. Experience-driven workflow induction~\parencite{awm} also suggests that recurring corrective sequences may be abstracted into reusable procedures, provided their applicability and actual effects are evaluated.

Co-evolution means that changes in one capability alter the behaviour and data distribution of the others. Improving spatial grounding may change where the controller attempts grasps; improving a controller may make formerly infeasible task plans useful. In turn, execution results can expose the deficiencies of both. A shared improvement process should propose targeted changes, evaluate their effects on the integrated system and retain the option of rolling them back. Model updates, procedural edits, newly composed tools and memory retrieval can all be appropriate; it is an empirical question which mechanism is most economical for a given failure.

\section{End-to-end evaluation and future directions}
\label{sec:vnext}

Scenario-driven development is central to RoboNoid. Tasks such as bringing an object to a person, operating an unfamiliar appliance, or seeking information from a person require coordinated perception, planning, manipulation and locomotion. The challenge is not only to demonstrate each behaviour, but to establish whether the agent can preserve the task's intended outcome under changes in observations and execution. In particular, task progress is not necessarily monotonic: a cup that was successfully acquired may later be dropped. Verification should therefore distinguish achieved effects from completed commands, and task progress from the agent's own belief about it. Reporting successful completion, recovery, time, interventions and resource costs together can reveal different limitations of the system. Long-horizon mobile manipulation and whole-body demonstrations~\parencite{mobilealoha,helix02} are useful precedents for evaluating coordinated behaviour beyond isolated manipulation skills.

\begin{figure*}[!t]
    \centering
    \includegraphics[width=.86\textwidth]{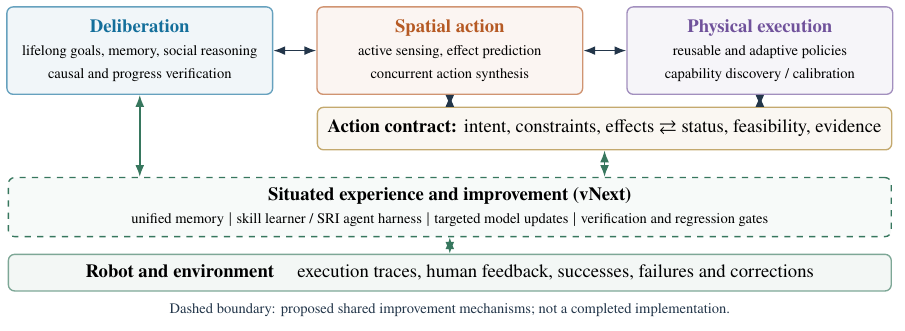}
    \caption{RoboNoid vNext: an architectural direction towards improvement through shared experience and end-to-end evaluation. The improvement mechanism is cross-cutting rather than a fourth control layer. The diagram includes proposed capabilities.}
    \label{fig:vnext}
\end{figure*}

Figure~\ref{fig:vnext} outlines directions towards richer world modelling, memory for longer-horizon tasks and experience-driven improvement. At the task level, causal reasoning about physical and interface interactions, goal tracking and outcome verification are important next steps. Spatial action planning can benefit from broader context, including maps and persistent memory. At the execution level, more closely integrated whole-body motion and manipulation, including vision-centric control policies and learned representations of physical intent, may reduce limitations of the current skill interfaces. More unified models can be introduced when supported by evidence, without committing in advance to a particular partition of capabilities.

A further objective is to use situated experience to improve the agent over time. This may include changes to retrieved knowledge, procedures, grounding models, execution policies or their interfaces. Changes should be evaluated for their effects on the integrated system and on previously acquired capabilities. Fast control and safety-critical feedback must remain protected from unvalidated updates. These are research directions, rather than a claim that the current system already performs autonomous lifelong learning.

\section{Related work and positioning}

\paragraph{Foundation models and grounded agents} Early work on language-conditioned robotic agents combines semantic planning with estimates of available robot affordances, as in SayCan~\parencite{saycan}. Inner Monologue~\parencite{inner} incorporates environmental feedback, including success detection and scene descriptions, into subsequent planning. PaLM-E~\parencite{palme} learns a multimodal model for embodied reasoning by integrating visual and state information into a language model. ReAct~\parencite{react} studies the interplay of reasoning, action and observation in interactive environments, while Code as Policies~\parencite{cap} shows how language models can synthesize programs that compose existing robot capabilities. RoboNoid shares this emphasis on grounded decisions and feedback, but focuses on the interaction between task-level reasoning, spatial action planning and physical realization in humanoids. This distinction is motivated by the need to coordinate balance, viewpoint, manipulation and locomotion, rather than simply select a sequence of available skills.

\paragraph{End-to-end robot learning and generalist policies} RT-2, OpenVLA and $\pi_{0.5}$ investigate how visual-language priors and diverse robot data can support generalizable action generation~\parencite{rt2,openvla,pi05}. Open X-Embodiment~\parencite{openx} establishes a precedent for sharing datasets and learning across heterogeneous embodiments. GR00T N1~\parencite{grootn1} combines a vision-language component with a learned action-generation component for humanoid manipulation, and Gemini Robotics~\parencite{geminirobotics} develops generalist vision-language-action models for dexterous manipulation and embodied reasoning. Mobile ALOHA~\parencite{mobilealoha} demonstrates coordinated bimanual mobile manipulation learned from whole-body teleoperation. Figure AI's Helix 02 technical announcement~\parencite{helix02} describes integrated humanoid locomotion, manipulation and balance using capabilities at different timescales, including a learned whole-body controller. Its use of additional onboard cameras and tactile feedback represents a different sensing trade-off from our RGB-centric approach. These systems motivate tighter integration of action capabilities; their different tasks, embodiments and supervision make direct performance comparisons inappropriate here. In particular, end-to-end task execution does not require that every capability be trained jointly in a single model. RoboNoid treats the choice of capability boundaries, sensing requirements and data as subjects of investigation rather than fixed assumptions.

\paragraph{Spatial representations and physical realization} VLMaps~\parencite{vlmaps} connects geometric maps to language through visual-semantic features, and ConceptGraphs~\parencite{conceptgraphs} builds open-vocabulary object graphs for perception and planning. Such work motivates spatial memory representations that can support decisions beyond the current view. RANGER emphasizes the construction of usable spatial knowledge from monocular RGB and separately acquired visual experience. NaVILA~\parencite{navila} investigates vision-language-action modelling for legged navigation. LeVERB, HEAD and SONIC explore different interfaces to whole-body control, including learned motion instructions, hand and eye targets, and motion tracking with planning~\parencite{leverb,head,sonic}. These offer alternatives to fixed symbolic skills or direct joint commands. In our architecture, the relationship between spatial grounding and physical realization is bidirectional: feasibility and realized effects inform subsequent decisions, while the details of motion generation may change with the backend.

\paragraph{Learning from visual experience} R3M~\parencite{r3m} demonstrates how representations learned from human video can support downstream robot manipulation. Video-based policy generation~\parencite{universalvideo} and UniSim~\parencite{unisim} explore using generative models to predict or simulate visual interaction. Such approaches complement robot demonstrations and simulation, and motivate investigating which aspects of behaviour can be learned without robot-specific motor supervision. Our ExoActor investigation explores the conversion of imagined visual demonstrations into whole-body motion references; an open direction is using alternative imagined outcomes to supervise failure recognition and recovery. The usefulness of generated experience depends on its fidelity to physical constraints and the task, not only visual plausibility.

\paragraph{Agentic experience and memory} Reflexion~\parencite{reflexion} stores verbal reflections from previous attempts to improve subsequent decisions without parameter updates. ExpeL~\parencite{expel} extracts reusable insights from past trajectories and retrieves relevant experience at inference time. Voyager~\parencite{voyager} develops a persistent library of executable skills through exploration and feedback in Minecraft. Agent Workflow Memory~\parencite{awm} induces reusable workflows from web interaction trajectories. Together with Cradle~\parencite{cradle}, these approaches illustrate mechanisms for accumulating experience outside a model's parameters. MART and PRACTICE~\parencite{mart,practice} complement them by learning how to select useful multimodal trajectories and how to maintain procedural knowledge using success and failure. The challenge for humanoids is not just transferring such mechanisms to a new action space. A past experience may depend on a particular stance, object configuration or execution policy; the same correction may no longer be feasible in a changed environment. Evidence about actual outcomes, their causes and the applicability of reusable knowledge therefore matters. The shared state and memory in RoboNoid provide an architectural basis for investigating these dependencies, while model or skill changes remain subject to end-to-end evaluation.

\section{Discussion and limitations}

\paragraph{Why composition now?} The requirements of general humanoid intelligence are not known in advance. Training a single model to jointly solve every abstraction and physical control problem may impose expensive coupled supervision before we know what is actually required. Composable systems provide a way to investigate failures and generate relevant situated data. This is a pragmatic methodological choice, not a claim that modularity is inherently superior. Boundaries can be learned, consolidated or changed when evidence supports doing so. Their utility should be judged by end-to-end outcomes, acquisition and integration costs, and the ability to evolve capabilities.

\paragraph{The limits of RGB.} Vision cannot fully reveal contact forces, latent object state or robot dynamics. Monocular geometry may fail, and active perception adds time and can be constrained by motion and safety. Some tasks require other modalities, calibration or high precision. The question is not whether these are useful, but when their cost and dependence are necessary. This distinction is particularly relevant to sensor-rich whole-body approaches such as Helix 02~\parencite{helix02}. Safety-critical constraints cannot be relaxed merely to make data collection cheaper.

\paragraph{The limits of learning from experience.} An agent's own interpretation of a trajectory can be mistaken. Experience can reinforce incorrect causal explanations, induce regressions, or overfit a familiar environment. Reliable feedback, independent evaluation, provenance, retention and rollback are central to the research agenda. In physical settings, learning must not require unrestricted risky exploration. Social settings further demand consent and appropriate treatment of human data.

The report describes an evolving architecture, its reference implementations and complementary research outcomes. The examples and cited studies concern different configurations and tasks; they should not be read as a single evaluation of the latest integrated RoboNoid system. A central objective of continuing work is to establish how the different capabilities perform together in repeated, realistic tasks, and how observations of failure inform subsequent development.

\section{Conclusion}

RoboNoid treats the scalability of humanoid intelligence as a problem of acquiring useful data, composing capabilities, and adapting to the world in which those capabilities are exercised. RGB-centric learning reduces some constraints on acquisition; situated feedback reduces some assumptions about perception and execution; and composability lets us investigate and improve the agent without requiring all its capabilities to be relearned together. The proposed architecture makes this approach concrete, supporting end-to-end experimentation and providing a basis for the evolution of capabilities. Our complementary investigations validate aspects of the design, identify remaining gaps, and explore alternatives that can inform its next iteration. The next measure of progress is not simply more capable components, but what they enable an agent to accomplish together under realistic conditions, and how the resulting experience informs the next iteration of the system.

\printbibliography[title={References}]

@article{being0,
  title={Being-0: A Humanoid Robotic Agent with Vision-Language Models and Modular Skills},
  author={Yuan, Haoqi and Bai, Yu and Fu, Yuhui and Zhou, Bohan and Feng, Yicheng and Xu, Xinrun and Zhan, Yi and Karlsson, B{\"o}rje F. and Lu, Zongqing},
  journal={arXiv preprint arXiv:2503.12533},
  year={2025},
  url={https://arxiv.org/abs/2503.12533v2}
}

@inproceedings{egoactor,
  title={EgoActor: Grounding Task Planning into Spatial-aware Egocentric Actions for Humanoid Robots via Visual-Language Models},
  author={Bai, Yu and Yu, MingMing and Li, Chaojie and Bai, Ziyi and Wang, Xinlong and Karlsson, B{\"o}rje F.},
  year={2026},
  booktitle={Findings of the Association for Computational Linguistics: {EMNLP} 2026},
  month={10},
  address={Budapest, Hungary},
  publisher={Association for Computational Linguistics},
  pages={24059--24083},
  doi={10.18653/2026.findings-emnlp.1136}
}

@inproceedings{ranger,
  title={{RANGER}: A Monocular Zero-Shot Semantic Navigation Framework through Visual Contextual Adaptation},
  author={Yu, Ming-Ming and Chen, Yi and Karlsson, B{\"o}rje F. and Wu, Wenjun},
  booktitle={2026 IEEE International Conference on Robotics and Automation (ICRA)},
  year={2026},
  pages={11818--11825},
  publisher={IEEE},
  doi={10.1109/ICRA57385.2026.11696276}
}

@inproceedings{cradle,
  title = {{Cradle}: Empowering Foundation Agents towards General Computer Control},
  author = {Tan, Weihao and Zhang, Wentao and Xu, Xinrun and
            Xia, Haochong and Ding, Ziluo and Li, Boyu and
            Zhou, Bohan and Yue, Junpeng and Jiang, Jiechuan and
            Li, Yewen and An, Ruyi and Qin, Molei and
            Zong, Chuqiao and Zheng, Longtao and Wu, Yujie and
            Chai, Xiaoqiang and Bi, Yifei and Xie, Tianbao and
            Gu, Pengjie and Li, Xiyun and Zhang, Ceyao and
            Tian, Long and Wang, Chaojie and Wang, Xinrun and
            Karlsson, B{\"o}rje F. and An, Bo and
            Yan, Shuicheng and Lu, Zongqing},
  booktitle = {Proceedings of the 42nd International Conference on Machine Learning},
  series = {Proceedings of Machine Learning Research},
  volume = {267},
  pages = {58658--58725},
  publisher = {PMLR},
  year = {2025},
  url = {https://proceedings.mlr.press/v267/tan25h.html}
}

@inproceedings{switch,
  title={SWITCH: Benchmarking Modeling and Handling of Tangible Interfaces in Long-horizon Embodied Scenarios},
  author={Cheng, Juntao and Zhang, Wanyue and Yu, Zhiwei and Ren, Shuo and He, Zheqi and Xie, Shaoxuan and Yao, Guocai and Lin, Jieru and Karlsson, B{\"o}rje F. and Zhang, Jiajun},
  year={2026},
  booktitle={Findings of the Association for Computational Linguistics: {EMNLP} 2026},
  month={10},
  address={Budapest, Hungary},
  publisher={Association for Computational Linguistics},
  pages={23852--23875},
  doi={10.18653/2026.findings-emnlp.1127}
}

@inproceedings{mart,
  title={MLLM as Retriever: Interactively Learning Multimodal Retrieval for Embodied Agents},
  author={Yue, Junpeng and Xu, Xinrun and Karlsson, B{\"o}rje F. and Lu, Zongqing},
  booktitle={International Conference on Learning Representations},
  year={2025},
  url={https://proceedings.iclr.cc/paper_files/paper/2025/hash/4e447acb68f57e29234bc0eb19896f11-Abstract-Conference.html}
}

@inproceedings{proprio,
  title={Towards Proprioception-Aware Embodied Planning for Dual-Arm Humanoid Robots},
  author={Li, Boyu and He, Siyuan and Xu, Hang and Yuan, Haoqi and Zang, Yu and Hu, Liwei and Yue, Junpeng and Jiang, Zhenxiong and Hu, Pengbo and Karlsson, B{\"o}rje F. and Zhao, Dongbin and Tang, Yehui and Lu, Zongqing},
  booktitle={2026 IEEE International Conference on Robotics and Automation (ICRA)},
  year={2026},
  pages={13011--13018},
  publisher={IEEE},
  doi={10.1109/ICRA57385.2026.11696567}
}

@article{deepphy,
  title={DeepPHY: Benchmarking Agentic VLMs on Physical Reasoning},
  author={Xu, Xinrun and Bu, Pi and Wang, Ye and Karlsson, B{\"o}rje F. and Wang, Ziming and Song, Tengtao and Zhu, Qi and Song, Jun and Ding, Zhiming and Zheng, Bo},
  journal={Proceedings of the AAAI Conference on Artificial Intelligence},
  volume={40},
  number={40},
  pages={34160--34168},
  year={2026},
  doi={10.1609/aaai.v40i40.40711}
}

@inproceedings{xdiffvla,
  title={X-DiffVLA: X-Embodied Diffusion Action Heads for Vision-Language-Action Models},
  author={Li, Boyu and Xu, Chaoyi and Yuan, Haoqi and Xu, Xinrun and Karlsson, B{\"o}rje F. and Zhao, Dongbin and Li, Haoran and Lu, Zongqing},
  year={2026},
  booktitle={Proceedings of Robotics: Science and Systems},
  address={Sydney, Australia},
  month={7},
  doi={10.15607/RSS.2026.XXII.081}
}

@article{grootn1,
  title={GR00T N1: An Open Foundation Model for Generalist Humanoid Robots},
  author={{NVIDIA} and Bjorck, Johan and Casta{\~n}eda, Fernando and Cherniadev, Nikita and Da, Xingye and Ding, Runyu and Fan, Linxi "Jim" and Fang, Yu and Fox, Dieter and Hu, Fengyuan and Huang, Spencer and Jang, Joel and Jiang, Zhenyu and Kautz, Jan and Kundalia, Kaushil and Lao, Lawrence and Li, Zhiqi and Lin, Zongyu and Lin, Kevin and Liu, Guilin and Llontop, Edith and Magne, Loic and Mandlekar, Ajay and Narayan, Avnish and Nasiriany, Soroush and Reed, Scott and Tan, You Liang and Wang, Guanzhi and Wang, Zu and Wang, Jing and Wang, Qi and Xiang, Jiannan and Xie, Yuqi and Xu, Yinzhen and Xu, Zhenjia and Ye, Seonghyeon and Yu, Zhiding and Zhang, Ao and Zhang, Hao and Zhao, Yizhou and Zheng, Ruijie and Zhu, Yuke},
  journal={arXiv preprint arXiv:2503.14734},
  year={2025},
  url={https://arxiv.org/abs/2503.14734}
}

@article{leverb,
  title={LeVERB: Humanoid Whole-Body Control with Latent Vision-Language Instruction},
  author={Xue, Haoru and Huang, Xiaoyu and Niu, Dantong and Liao, Qiayuan and Kragerud, Thomas and Gravdahl, Jan Tommy and Peng, Xue Bin and Shi, Guanya and Darrell, Trevor and Sreenath, Koushil and Sastry, S. Shankar},
  journal={arXiv preprint arXiv:2506.13751},
  year={2025},
  url={https://arxiv.org/abs/2506.13751}
}

@inproceedings{head,
  title={Hand-Eye Autonomous Delivery: Learning Humanoid Navigation, Locomotion and Reaching},
  author={Chen, Sirui and Ye, Yufei and Cao, Zi-ang and Xu, Pei and Lew, Jennifer and Liu, Karen},
  year={2025},
  url={https://proceedings.mlr.press/v305/chen25e.html},
  booktitle={Proceedings of the 9th Conference on Robot Learning},
  series={Proceedings of Machine Learning Research},
  volume={305},
  pages={4058--4073},
  publisher={PMLR}
}

@article{sonic,
  title={SONIC: Supersizing Motion Tracking for Natural Humanoid Whole-Body Control},
  author={Luo, Zhengyi and Yuan, Ye and Wang, Tingwu and Li, Chenran and Casta{\~n}eda, Fernando and Chen, Sirui and Cao, Zi-Ang and Li, Jiefeng and Minor, David and Ben, Qingwei and Park, Jinhyung and Sami, David and Wang, Zi and Da, Xingye and Ding, Runyu and Hogg, Cyrus and Song, Lina and Lim, Edy and Jeong, Eugene and He, Tairan and Xue, Haoru and Xiao, Wenli and Yuen, Simon and Kautz, Jan and Chang, Yan and Iqbal, Umar and Fan, Linxi Jim and Zhu, Yuke},
  journal={Science Robotics},
  year={2026},
  volume={11},
  number={117},
  eid={eaed4592},
  doi={10.1126/scirobotics.aed4592}
}

@inproceedings{saycan,
  title={Do As I Can, Not As I Say: Grounding Language in Robotic Affordances},
  author={Ichter, Brian and Brohan, Anthony and Chebotar, Yevgen and Finn, Chelsea and Hausman, Karol and Herzog, Alexander and Ho, Daniel and Ibarz, Julian and Irpan, Alex and Jang, Eric and Julian, Ryan and Kalashnikov, Dmitry and Levine, Sergey and Lu, Yao and Parada, Carolina and Rao, Kanishka and Sermanet, Pierre and Toshev, Alexander T. and Vanhoucke, Vincent and Xia, Fei and Xiao, Ted and Xu, Peng and Yan, Mengyuan and Brown, Noah and Ahn, Michael and Cortes, Omar and Sievers, Nicolas and Tan, Clayton and Xu, Sichun and Reyes, Diego and Rettinghouse, Jarek and Quiambao, Jornell and Pastor, Peter and Luu, Linda and Lee, Kuang-Huei and Kuang, Yuheng and Jesmonth, Sally and Joshi, Nikhil J. and Jeffrey, Kyle and Ruano, Rosario Jauregui and Hsu, Jasmine and Gopalakrishnan, Keerthana and David, Byron and Zeng, Andy and Fu, Chuyuan Kelly},
  booktitle={Proceedings of The 6th Conference on Robot Learning},
  series={Proceedings of Machine Learning Research},
  volume={205},
  pages={287--318},
  year={2023},
  url={https://proceedings.mlr.press/v205/ichter23a.html},
  publisher={PMLR}
}

@inproceedings{rt2,
  title={RT-2: Vision-Language-Action Models Transfer Web Knowledge to Robotic Control},
  author={Zitkovich, Brianna and Yu, Tianhe and Xu, Sichun and Xu, Peng and Xiao, Ted and Xia, Fei and Wu, Jialin and Wohlhart, Paul and Welker, Stefan and Wahid, Ayzaan and Vuong, Quan and Vanhoucke, Vincent and Tran, Huong and Soricut, Radu and Singh, Anikait and Singh, Jaspiar and Sermanet, Pierre and Sanketi, Pannag R. and Salazar, Grecia and Ryoo, Michael S. and Reymann, Krista and Rao, Kanishka and Pertsch, Karl and Mordatch, Igor and Michalewski, Henryk and Lu, Yao and Levine, Sergey and Lee, Lisa and Lee, Tsang-Wei Edward and Leal, Isabel and Kuang, Yuheng and Kalashnikov, Dmitry and Julian, Ryan and Joshi, Nikhil J. and Irpan, Alex and Ichter, Brian and Hsu, Jasmine and Herzog, Alexander and Hausman, Karol and Gopalakrishnan, Keerthana and Fu, Chuyuan and Florence, Pete and Finn, Chelsea and Dubey, Kumar Avinava and Driess, Danny and Ding, Tianli and Choromanski, Krzysztof Marcin and Chen, Xi and Chebotar, Yevgen and Carbajal, Justice and Brown, Noah and Brohan, Anthony and Arenas, Montserrat Gonzalez and Han, Kehang},
  booktitle={Proceedings of The 7th Conference on Robot Learning},
  series={Proceedings of Machine Learning Research},
  volume={229},
  pages={2165--2183},
  year={2023},
  url={https://proceedings.mlr.press/v229/zitkovich23a.html},
  publisher={PMLR}
}

@inproceedings{openvla,
  title={OpenVLA: An Open-Source Vision-Language-Action Model},
  author={Kim, Moo Jin and Pertsch, Karl and Karamcheti, Siddharth and Xiao, Ted and Balakrishna, Ashwin and Nair, Suraj and Rafailov, Rafael and Foster, Ethan P. and Lam, Grace and Sanketi, Pannag R. and Vuong, Quan and Kollar, Thomas and Burchfiel, Benjamin and Tedrake, Russ and Sadigh, Dorsa and Levine, Sergey and Liang, Percy and Finn, Chelsea},
  year={2025},
  url={https://proceedings.mlr.press/v270/kim25c.html},
  booktitle={Proceedings of the 8th Conference on Robot Learning},
  series={Proceedings of Machine Learning Research},
  volume={270},
  pages={2679--2713},
  publisher={PMLR}
}

@inproceedings{pi05,
  title={$\pi_{0.5}$: A Vision-Language-Action Model with Open-World Generalization},
  author={Black, Kevin and Brown, Noah and Darpinian, James and Dhabalia, Karan and Driess, Danny and Esmail, Adnan and Equi, Michael Robert and Finn, Chelsea and Fusai, Niccolo and Galliker, Manuel Y. and Ghosh, Dibya and Groom, Lachy and Hausman, Karol and Ichter, Brian and Jakubczak, Szymon and Jones, Tim and Ke, Liyiming and LeBlanc, Devin and Levine, Sergey and Li-Bell, Adrian and Mothukuri, Mohith and Nair, Suraj and Pertsch, Karl and Ren, Allen Z. and Shi, Lucy Xiaoyang and Smith, Laura and Springenberg, Jost Tobias and Stachowicz, Kyle and Tanner, James and Vuong, Quan and Walke, Homer and Walling, Anna and Wang, Haohuan and Yu, Lili and Zhilinsky, Ury},
  year={2025},
  url={https://proceedings.mlr.press/v305/black25a.html},
  booktitle={Proceedings of the 9th Conference on Robot Learning},
  series={Proceedings of Machine Learning Research},
  volume={305},
  pages={17--40},
  publisher={PMLR}
}

@inproceedings{navila,
  title={NaVILA: Legged Robot Vision-Language-Action Model for Navigation},
  author={Cheng, An-Chieh and Ji, Yandong and Yang, Zhaojing and Gongye, Zaitian and Zou, Xueyan and Kautz, Jan and Biyik, Erdem and Yin, Hongxu and Liu, Sifei and Wang, Xiaolong},
  year={2025},
  booktitle={Proceedings of Robotics: Science and Systems},
  address={Los Angeles, CA, USA},
  month={6},
  doi={10.15607/RSS.2025.XXI.018}
}

@article{practice,
  title={{PRACTICE}: From Experience to Expertise in Self-Evolving Embodied Agents},
  author={Bai, Ziyi and Li, Siqi and Huang, Tinglei and Karlsson, B{\"o}rje F.},
  journal={arXiv preprint arXiv:2608.30760},
  year={2026},
  url={https://arxiv.org/abs/2608.30760}
}

@article{embmem,
  title={When the Past Shapes Action: Understanding Embodied Memory},
  author={Li, Siqi and Bai, Ziyi and Xu, Xinrun and Huang, Tinglei and Karlsson, B{\"o}rje F.},
  year={2026},
  journal={arXiv preprint arXiv:2610.41688},
  url={https://arxiv.org/abs/2610.41688}
}

@article{viloman,
  author = {Tian, Zejie and Hou, Ruibing and Ma, Bingpeng and Karlsson, B{\"o}rje F. and Shan, Shiguang},
  title = {{ViLoMan}: Learning Visual--Proprioceptive Whole-Body Loco-Manipulation Skills for Humanoid Robots},
  journal = {arXiv preprint arXiv:2609.19340},
  year = {2026},
  url = {https://arxiv.org/abs/2609.19340}
}

@article{exoactor,
  author = {Zhou, Yanghao and Ma, Jingyu and Peng, Yibo and Sun, Zhenguo and Bai, Yu and Karlsson, B{\"o}rje F.},
  title = {{ExoActor}: Exocentric Video Generation as Generalizable Interactive Humanoid Control},
  journal = {arXiv preprint arXiv:2604.27711},
  year = {2026},
  url = {https://arxiv.org/abs/2604.27711}
}

@unpublished{muris,
  title={{MURIS}: A Map-Centric Multimodal Model and Benchmark for Indoor Spatial Reasoning},
  author={Yu, Ming-Ming and Bai, Yu and Mei, Yanghong and Bai, Ziyi and Chen, Yi and Zhao, Guiyu and Karlsson, B{\"o}rje F. and Wu, Wenjun},
  year={2026},
  note={Manuscript}
}

@article{react,
  title={{ReAct}: Synergizing Reasoning and Acting in Language Models},
  author={Shunyu Yao and Jeffrey Zhao and Dian Yu and Nan Du and Izhak Shafran and Karthik Narasimhan and Yuan Cao},
  year={2023},
  journal={arXiv preprint arXiv:2210.03629},
  url={https://arxiv.org/abs/2210.03629}
}

@article{inner,
  title={Inner Monologue: Embodied Reasoning through Planning with Language Models},
  author={Wenlong Huang and Fei Xia and Ted Xiao and Harris Chan and Jacky Liang and Pete Florence and Andy Zeng and Jonathan Tompson and Igor Mordatch and Yevgen Chebotar and Pierre Sermanet and Noah Brown and Tomas Jackson and Linda Luu and Sergey Levine and Karol Hausman and Brian Ichter},
  year={2022},
  journal={arXiv preprint arXiv:2207.05608},
  url={https://arxiv.org/abs/2207.05608}
}

@article{reflexion,
  title={{Reflexion}: Language Agents with Verbal Reinforcement Learning},
  author={Noah Shinn and Federico Cassano and Edward Berman and Ashwin Gopinath and Karthik Narasimhan and Shunyu Yao},
  year={2023},
  journal={arXiv preprint arXiv:2303.11366},
  url={https://arxiv.org/abs/2303.11366}
}

@article{expel,
  title={{ExpeL}: {LLM} Agents Are Experiential Learners},
  author={Andrew Zhao and Daniel Huang and Quentin Xu and Matthieu Lin and Yong-Jin Liu and Gao Huang},
  year={2023},
  journal={arXiv preprint arXiv:2308.10144},
  url={https://arxiv.org/abs/2308.10144}
}

@article{voyager,
  title={{Voyager}: An Open-Ended Embodied Agent with Large Language Models},
  author={Guanzhi Wang and Yuqi Xie and Yunfan Jiang and Ajay Mandlekar and Chaowei Xiao and Yuke Zhu and Linxi Fan and Anima Anandkumar},
  year={2023},
  journal={arXiv preprint arXiv:2305.16291},
  url={https://arxiv.org/abs/2305.16291}
}

@article{awm,
  title={Agent Workflow Memory},
  author={Zora Zhiruo Wang and Jiayuan Mao and Daniel Fried and Graham Neubig},
  year={2024},
  journal={arXiv preprint arXiv:2409.07429},
  url={https://arxiv.org/abs/2409.07429}
}

@article{conceptgraphs,
  title={{ConceptGraphs}: Open-Vocabulary {3D} Scene Graphs for Perception and Planning},
  author={Qiao Gu and Alihusein Kuwajerwala and Sacha Morin and Krishna Murthy Jatavallabhula and Bipasha Sen and Aditya Agarwal and Corban Rivera and William Paul and Kirsty Ellis and Rama Chellappa and Chuang Gan and Celso Miguel de Melo and Joshua B. Tenenbaum and Antonio Torralba and Florian Shkurti and Liam Paull},
  year={2023},
  journal={arXiv preprint arXiv:2309.16650},
  url={https://arxiv.org/abs/2309.16650}
}

@article{vlmaps,
  title={Visual Language Maps for Robot Navigation},
  author={Chenguang Huang and Oier Mees and Andy Zeng and Wolfram Burgard},
  year={2022},
  journal={arXiv preprint arXiv:2210.05714},
  url={https://arxiv.org/abs/2210.05714}
}

@article{mobilealoha,
  title={{Mobile ALOHA}: Learning Bimanual Mobile Manipulation with Low-Cost Whole-Body Teleoperation},
  author={Zipeng Fu and Tony Z. Zhao and Chelsea Finn},
  year={2024},
  journal={arXiv preprint arXiv:2401.02117},
  url={https://arxiv.org/abs/2401.02117}
}

@article{cap,
  title={Code as Policies: Language Model Programs for Embodied Control},
  author={Jacky Liang and Wenlong Huang and Fei Xia and Peng Xu and Karol Hausman and Brian Ichter and Pete Florence and Andy Zeng},
  year={2022},
  journal={arXiv preprint arXiv:2209.07753},
  url={https://arxiv.org/abs/2209.07753}
}

@article{r3m,
  title={{R3M}: A Universal Visual Representation for Robot Manipulation},
  author={Suraj Nair and Aravind Rajeswaran and Vikash Kumar and Chelsea Finn and Abhinav Gupta},
  year={2022},
  journal={arXiv preprint arXiv:2203.12601},
  url={https://arxiv.org/abs/2203.12601}
}

@article{unisim,
  title={Learning Interactive Real-World Simulators},
  author={Sherry Yang and Yilun Du and Kamyar Ghasemipour and Jonathan Tompson and Leslie Kaelbling and Dale Schuurmans and Pieter Abbeel},
  year={2023},
  journal={arXiv preprint arXiv:2310.06114},
  url={https://arxiv.org/abs/2310.06114}
}

@article{universalvideo,
  title={Learning Universal Policies via Text-Guided Video Generation},
  author={Yilun Du and Mengjiao Yang and Bo Dai and Hanjun Dai and Ofir Nachum and Joshua B. Tenenbaum and Dale Schuurmans and Pieter Abbeel},
  year={2023},
  journal={arXiv preprint arXiv:2302.00111},
  url={https://arxiv.org/abs/2302.00111}
}

@article{palme,
  title={{PaLM-E}: An Embodied Multimodal Language Model},
  author={Danny Driess and Fei Xia and Mehdi S. M. Sajjadi and Corey Lynch and Aakanksha Chowdhery and Brian Ichter and Ayzaan Wahid and Jonathan Tompson and Quan Vuong and Tianhe Yu and Wenlong Huang and Yevgen Chebotar and Pierre Sermanet and Daniel Duckworth and Sergey Levine and Vincent Vanhoucke and Karol Hausman and Marc Toussaint and Klaus Greff and Andy Zeng and Igor Mordatch and Pete Florence},
  year={2023},
  journal={arXiv preprint arXiv:2303.03378},
  url={https://arxiv.org/abs/2303.03378}
}

@article{geminirobotics,
  title={{Gemini Robotics}: Bringing {AI} into the Physical World},
  author={{Gemini Robotics Team} and others},
  year={2025},
  journal={arXiv preprint arXiv:2503.20020},
  url={https://arxiv.org/abs/2503.20020}
}

@article{openx,
  title={Open {X-Embodiment}: Robotic Learning Datasets and {RT-X} Models},
  author={{Open X-Embodiment Collaboration}},
  year={2023},
  journal={arXiv preprint arXiv:2310.08864},
  url={https://arxiv.org/abs/2310.08864}
}

@online{helix02,
  title={Introducing {Helix 02}: Full-Body Autonomy},
  author={{Figure AI}},
  year={2026},
  url={https://www.figure.ai/news/helix-02}
}

\end{document}